\documentclass[10pt,twocolumn,letterpaper]{article}
\usepackage[pagenumbers]{cvpr}
\usepackage{colortbl}
\usepackage{multirow}
\newcolumntype{C}[1]{>{\centering\arraybackslash}m{#1}}

\usepackage{stfloats}

\makeatletter
\@ifundefined{@LN@amsmath@patch}{}{%
  \let\linenomathpatch\@gobble
  \let\linenomathpatchAMS\@gobble
}
\makeatother

\ifdefined\pdfminorversion
\fi

\usepackage{algorithm,algpseudocode}
\definecolor{cvprblue}{rgb}{0.21,0.49,0.74}
\usepackage[pagebackref,breaklinks,colorlinks,allcolors=cvprblue]{hyperref}
\hypersetup{pdfauthor={Yuhao Fang, Chenglin Lu, Hefei Ling, Chun Lou},pdftitle={ThermoSight: From Appearance to Internal Thermofluid Fields via Physics-Guided Neural Reconstruction}}
\def\confName{CVPR}
\def\confYear{2027}

\title{ThermoSight: From Appearance to Internal Thermofluid Fields via Physics-Guided Neural Reconstruction}

\author{%
Yuhao Fang\textsuperscript{1}\quad
Chenglin Lu\textsuperscript{1}\quad
Hefei Ling\textsuperscript{2}\quad
Chun Lou\textsuperscript{1,*}\\[0.5em]
\parbox{0.96\textwidth}{\centering\normalsize
\textsuperscript{1}State Key Laboratory of Coal Combustion, Huazhong University of Science and Technology\\[0.3em]
\textsuperscript{2}School of Computer Science and Technology, Huazhong University of Science and Technology\\[0.3em]
{\small\ttfamily \{fyh2000520, m202571368, lhefei, clou\}@hust.edu.cn}}%
}

\begin{document}

\maketitle
\setlength{\parskip}{0pt}
\begin{abstract}
Most neural reconstruction methods focus on reproducing scene appearance, yet inferring physical states from visual observations is essential for machine understanding of the physical world. Reconstructing the internal temperature and soot fields of radiating thermofluids such as flames is challenging yet essential for improving energy efficiency and assessing heat exposure and fire spread. In flames, temperature and soot concentration jointly govern visible emission, and soot self-absorption alters the light that reaches each camera. With limited views, changes in the two fields can offset one another in the recorded RGB images, making temperature and soot difficult to disentangle. ThermoSight, a physics-guided neural framework for scientific imaging, jointly reconstructs continuous three-dimensional temperature and soot volume fraction fields from limited calibrated flame images without field-level ground truth during optimization. A differentiable spectral model links these fields to measured RGB through thermal emission, self-absorption, and calibrated camera responses. Temperature-dependent emission spectra and wavelength-dependent attenuation constrain shared fields across RGB channels, reducing temperature--soot ambiguity. Chromaticity consistency, observation-derived spatial constraints, and soot regularization complement intensity fitting. In a synthetic ablation, wavelength-resolved attenuation reduces temperature RMSE in the hot, sooty region from 234 to 112 K relative to a three-channel approximation, with nearly identical RGB PSNR (65.23 vs. 65.24 dB). Measured data yield coherent fields and competitive RGB reconstructions. These results demonstrate that physics-guided neural fields recover interpretable internal physical quantities beyond visually consistent appearance.
\par
\end{abstract}

\begin{figure}[t]
    \centering
    \includegraphics[width=0.88\columnwidth]{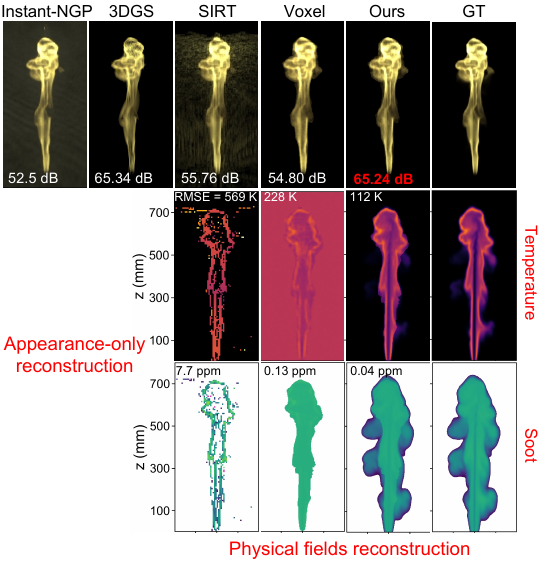}
    \caption{\textbf{From appearance to internal physical fields.} Synthetic RGB (top; PSNR in dB) and temperature/soot slices (middle/bottom). Field RMSE uses the hot, sooty volume (K for temperature; ppm for soot). Instant-NGP and 3DGS recover appearance; emission-only SIRT yields fragmented fields, while emission--absorption Voxel retains elevated background temperatures and nearly uniform soot. ThermoSight better matches the reference fields.}
    \label{fig:teaser}
\end{figure}

\flushbottom
\setlength{\parskip}{0pt plus 1pt}
\section{Introduction}
\label{sec:intro}

Established scene reconstruction methods (e.g., NeRF~\cite{mildenhall2020nerf} and 3D Gaussian Splatting [3DGS]~\cite{kerbl2023gaussian}) focus on reproducing scene appearance for novel-view synthesis. Inferring physical states from images is central to machine understanding of the physical world~\cite{lake2017building}. For a luminous flame observed from a few views, predicting its RGB appearance means matching the light that reaches the cameras. Recovering the internal fields also requires explaining what produced that light along each line of sight~\cite{grauer2023volumetric}. Cold smoke is a non-emitting flow, and its visible patterns provide cues to smoke density~\cite{chu2022physics,yu2023hyfluid}. A luminous flame is a self-emitting flow in which hot soot emits intense visible thermal radiation and absorbs some of the light emitted deeper in the flame~\cite{liu2013soot,huang2023hyperspectral}. A bright pixel can come from hotter soot or from more soot emitting along the viewing ray. Different three-dimensional temperature and soot fields can thus produce nearly identical RGB images~\cite{yuan2022endoscopic}. Coupled emission and absorption therefore make the internal fields of self-emitting flows difficult to recover from sparse RGB views. Yet recovering these fields can guide combustion optimization and particulate-emission control in gas turbines and furnaces~\cite{terlouw2025hydrogen,farooq2022laser,alden2023spatially,moore2017biofuel}, and inform assessments of heat exposure and fire spread~\cite{chen2023pool,finney2015buoyant}. Addressing this ambiguity calls for an image-formation model that links temperature and soot to the camera signal, together with constraints on the possible solutions~\cite{arridge2019solving,zha2022naf,feng2026pidef}.

We present ThermoSight, a physics-guided neural framework extending appearance-oriented reconstruction toward physical-field inference for scientific imaging. Three-dimensional temperature and soot volume fraction are represented as continuous neural fields~\cite{mildenhall2020nerf,mueller2022instant} and inferred from limited calibrated flame images through a differentiable spectral emission--absorption model~\cite{liu2013soot}. Planck's law sets the temperature dependence of soot emission. After wavelength-dependent self-absorption, the spectrum is integrated against calibrated camera responses~\cite{kuhn2011soot} so the same fields must explain all three RGB channels. Chromaticity consistency, observation-derived spatial constraints, and soot regularization complement intensity fitting to further reduce reconstruction ambiguity.

The synthetic comparison in \cref{fig:teaser} shows why image agreement alone is insufficient. Appearance-based methods can reproduce flame images~\cite{hasinoff2007photo,nazarenus2026gaussians}, and baselines that explicitly estimate physical fields can still yield large errors within the flame or its surroundings. The emission-only baseline produces fragmented fields, while the voxel baseline with emission and absorption retains an elevated background temperature and nearly uniform soot. ThermoSight achieves RGB PSNR comparable to 3DGS while recovering temperature and soot fields closer to the synthetic reference. Measured data, sparse-view tests, and ablations assess applicability, view dependence, and component effects.

Our main contributions are:
\begin{itemize}
    \item A physics-guided neural framework accurately reconstructs continuous 3D temperature and soot volume fraction fields from limited calibrated views, supporting arbitrary spatial queries and flexible sampling densities. This extends appearance-based reconstruction to scientific imaging of underlying physical states.

    \item A differentiable spectral flame-imaging model is developed to link temperature and soot volume fraction fields to calibrated RGB observations. Temperature-dependent emission spectra and wavelength-dependent self-absorption constrain temperature--soot ambiguity, complemented by chromaticity consistency, observation-derived spatial constraints, and soot regularization.

    \item We use synthetic flame data to quantify physical-field accuracy and measured flame images to assess reconstruction under laboratory conditions.
\end{itemize}

\section{Related Work}
\label{sec:formatting}

\noindent\textbf{Physics-inspired 3D Vision and Imaging.}
Explicit image formation helps separate physical quantities that contribute to the same image. In inverse rendering, light transport constrains the decomposition of geometry, reflectance, and illumination~\cite{jin2023tensoir,wu2025pbrnerf}. In thermal imaging, radiometric drift complicates multi-view consistency~\cite{aydin2026thermal}, while emission and reflection contribute to the same thermal signal. PhysIR-Splat models both components in the renderer~\cite{gao2026physir}, and dual-band video provides spectral and temporal cues to separate them~\cite{narayanan2026dualband}. Physical imaging models can also probe quantities inside a volume, as NeST demonstrates by recovering stress from polarization measurements via photoelasticity~\cite{dave2025nest}. ThermoSight combines Planck's law with spectral emission--absorption transport to link internal temperature and soot fields to calibrated RGB observations.

\noindent\textbf{Neural Fields and Fluid Reconstruction.}
Continuous fields make it possible to fit a volume through a differentiable measurement model~\cite{mildenhall2020nerf,sitzmann2020implicit,tancik2020fourier,mueller2022instant,chen2022tensorf}. The measurement model ties X-ray projections to attenuation fields represented by neural networks~\cite{rueckert2022neat,zha2022naf,cai2024saxnerf} or Gaussians~\cite{zha2024r2gaussian,wu2025dgr}, and schlieren images to refractive-index variations~\cite{lu2026refractive,he2026tensorbost}. Measurement-domain fitting and prior images offer additional constraints when observations are sparse~\cite{sun2021coil,shen2024nerp}. For non-emitting smoke, visible patterns constrain density, and scalar transport links successive frames~\cite{gregson2014capture,eckert2018coupled,eckert2019scalarflow,zang2020tomofluid}. Governing equations link scalar observations to fluid dynamics~\cite{raissi2020hidden,chu2022physics,yu2023hyfluid}, and explicit transport models promote temporal consistency~\cite{wang2024physics,qiu2024neusmoke,zhang2026gausmoke}. Learned priors and synthesized views address sparse camera coverage~\cite{franz2021global,franz2023learning,gao2025fluidnexus,li2026smokesvd}. In self-emitting flames, hot soot both emits and absorbs light. Different temperature and soot fields can thus produce similar RGB images across simultaneous views.

\noindent\textbf{Image-Based Flame Diagnostics.}
Recovering a flame's volume from sparse views calls for spatial or temporal structure~\cite{ihrke2004image,hasinoff2007photo,okabe2015fluid,nazarenus2026gaussians}, while radiative transport governs how that volume appears in the images~\cite{7940071}. To recover physical quantities, pyrometry exploits the spectral dependence of soot emission~\cite{snelling2002spectrally,kuhn2011soot}, and tomographic, light-field, or hyperspectral measurements provide additional angular or spectral information~\cite{huang2009reconstruction,huang2012simultaneous,yu2021temperature,qi2023simultaneous,huang2023hyperspectral}. Because intervening soot can reabsorb emission, the measured spectrum depends on the distribution of temperature and soot along the ray~\cite{liu2013soot,kempema2018effect,yuan2022endoscopic}. ThermoSight models this spectral coupling to jointly recover continuous temperature and soot fields from calibrated multi-view RGB.

\setlength{\parskip}{0pt}
\section{Method}
\label{sec:method}

\subsection{Overview}
\label{sec:method_overview}

\begin{figure*}[tbp]
    \centering
    \includegraphics[width=\textwidth]{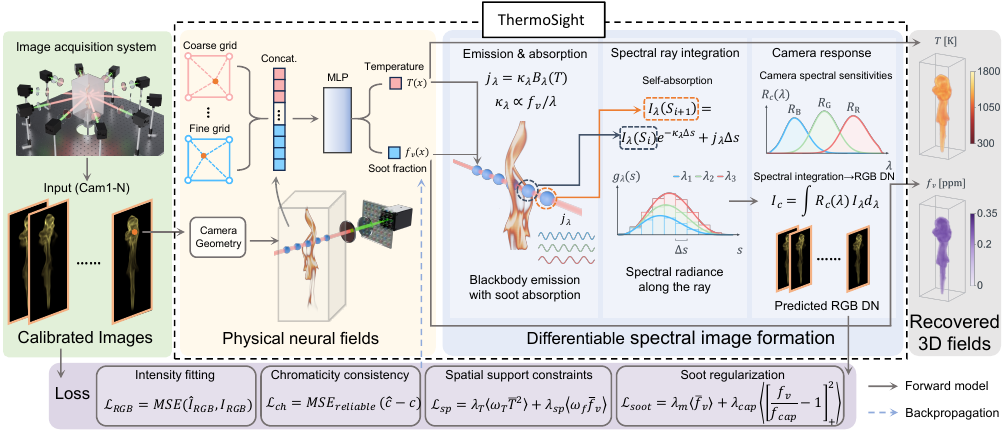}
    \caption{\textbf{Overview of ThermoSight.} Calibrated multi-view images supervise continuous temperature and soot fields through differentiable spectral rendering and four complementary objectives. The ray-update sketch uses a thin-segment approximation; reconstruction uses the analytic solution in \cref{eq:segment_rendering}.}
    \label{fig:thermosight}
\end{figure*}

ThermoSight jointly reconstructs continuous temperature $T(\mathbf x)$ and soot volume fraction $f_v(\mathbf x)$ in a bounded three-dimensional domain $\Omega$ from calibrated multi-view flame images $\{\mathbf I_k\}_{k=1}^{N_c}$ at a single instant. As shown in \cref{fig:thermosight}, a hash encoder and shared multilayer perceptron (MLP) represent both fields (\cref{sec:physical_fields}). Differentiable spectral rendering models thermal emission, self-absorption, and camera response (\cref{sec:spectral_formation}); intensity, chromaticity, spatial, and soot constraints guide optimization without field-level ground truth (\cref{sec:optimization}).

\subsection{Physical Neural Fields}
\label{sec:physical_fields}

World coordinates are normalized to $\widetilde{\mathbf x}\in[0,1]^3$ within $\Omega$. A multiresolution hash encoder~\cite{mueller2022instant} concatenates interpolated trainable features across resolutions; a shared MLP predicts two scalar outputs:
\begin{equation}
    (a_\theta,b_\theta)(\mathbf x)
    = F_\theta\!\left(\bigoplus_{\ell=1}^{L}
      H_\ell(\widetilde{\mathbf x})\right),
    \label{eq:field_encoding}
\end{equation}
where $H_\ell$ is the interpolated feature at level $\ell$, $\bigoplus$ denotes concatenation, and $\theta$ includes both encoder and MLP parameters. The outputs are mapped to admissible physical quantities by
\begin{equation}
\begin{aligned}
    T_\theta(\mathbf x)
      &= T_{\min}+(T_{\max}-T_{\min})\sigma(a_\theta),\\
    f_{v,\theta}(\mathbf x)
      &= f_{\mathrm{scale}}\,\operatorname{softplus}(b_\theta),
\end{aligned}
\label{eq:physical_outputs}
\end{equation}
with sigmoid $\sigma$ and a positive soot scale $f_{\mathrm{scale}}$. Temperature is expressed in kelvin and $f_v$ is dimensionless; $f_{\mathrm{scale}}$ is a numerical scale rather than an upper concentration bound. A fixed exterior mask can exclude soot as described in \cref{sec:spatial_support}; subsequent equations use the effective fields after this exclusion.

\subsection{Differentiable Spectral Image Formation}
\label{sec:spectral_formation}

\noindent\textbf{Thermal emission and absorption.}
The forward model describes thermal emission from soot under local thermal equilibrium, with negligible scattering. Gas-band emission and chemiluminescence are outside this model. Following soot-emission diagnostics~\cite{snelling2002spectrally,liu2013soot}, the local absorption and emission coefficients are
\begin{equation}
\begin{aligned}
    \kappa_\lambda(\mathbf x)
      &= \frac{6\pi E(\lambda)}{\lambda}\,f_v(\mathbf x),\\
    j_\lambda(\mathbf x)
      &= \kappa_\lambda(\mathbf x)B_\lambda(T(\mathbf x)),
\end{aligned}
\label{eq:soot_coefficients}
\end{equation}
where $E(\lambda)$ is the prescribed soot absorption function and
\begin{equation}
    B_\lambda(T)=\frac{2hc^2}{\lambda^5}
    \left[\exp\!\left(\frac{hc}{\lambda k_B T}\right)-1\right]^{-1}
    \label{eq:planck_radiance}
\end{equation}
is Planck's spectral radiance. Here, $h$, $c$, and $k_B$ denote Planck's constant, the speed of light, and Boltzmann's constant. Wavelengths and ray lengths are expressed in metres. For synthetic and measured data, we fix $E(\lambda)=0.26$ rather than estimating it jointly with $f_v$.

\noindent\textbf{Spectral ray integration.}
For pixel $p$ of camera $k$, the calibrated ray is $\mathbf r(s)=\mathbf o_k+s\mathbf d_{k,p}$, with unit direction $\mathbf d_{k,p}$ and intersections $s_n,s_f$ with $\Omega$. The spectral radiance reaching the camera is
\begin{equation}
\begin{aligned}
    \widehat L_\lambda(\mathbf r)
      &=\int_{s_n}^{s_f}e^{-\tau_\lambda(s)}
        j_\lambda(\mathbf r(s))\,\mathrm ds,\\
    \tau_\lambda(s)
      &=\int_{s_n}^{s}\kappa_\lambda(\mathbf r(u))\,\mathrm du,
\end{aligned}
\label{eq:spectral_transport}
\end{equation}
with zero incident radiance at the far boundary. Thus, emission from each point is attenuated by the soot between that point and the camera. For piecewise-constant ray segments ordered from near to far, the integral is evaluated analytically as
\begin{equation}
    \widehat L_\lambda(\mathbf r)
    =\sum_{i=1}^{N_s}\mathcal T_{\lambda,i}
      \left(1-e^{-\kappa_{\lambda,i}\Delta s_i}\right)
      B_\lambda(T_i),
    \label{eq:segment_rendering}
\end{equation}
where $\mathcal T_{\lambda,i}=\exp(-\sum_{j<i}\kappa_{\lambda,j}\Delta s_j)$ and $\Delta s_i$ is the segment length. The source term is differentiable at zero soot and approaches $j_{\lambda,i}\Delta s_i$ for optically thin segments (\cref{fig:thermosight}). See Supplementary Sec.~B.3 for the derivation and chromaticity analysis.

\noindent\textbf{Camera response and radiometric calibration.}
Spectral radiance is integrated against each channel's effective response before conversion to digital numbers (DN):
\begin{equation}
\begin{aligned}
    C_{k,p,c}
      &=\int_{\Lambda}R_{k,c}(\lambda)
        \widehat L_\lambda(\mathbf r_{k,p})\,\mathrm d\lambda,\\
    \widehat I_{k,p,c}
      &=d_{k,c}+g_{k,c}\!\left(t_k C_{k,p,c}\right),
\end{aligned}
\label{eq:camera_observation}
\end{equation}
where $c\in\{R,G,B\}$, $t_k$ is exposure time, and $d_{k,c}$ is the exposure-matched dark reference. The effective response $R_{k,c}$ includes spectral sensitivity, optical transmission, and photon conversion when sensitivity is specified as quantum efficiency. Explicit response characterization follows quantitative color-camera pyrometry~\cite{kuhn2011soot}. For our experimental cameras, $g_{k,c}$ is a fixed quadratic function fitted from blackbody calibration, with its input scale included in the fitted coefficients. Synthetic observations use a known linear gain. No display gamma or white balancing is applied.

\subsection{Constraints and Optimization}
\label{sec:optimization}

The parameters $\theta$ are optimized using
\begin{equation}
    \mathcal L=\mathcal L_{\mathrm{RGB}}
      +\lambda_{\mathrm{ch}}\mathcal L_{\mathrm{ch}}
      +\mathcal L_{\mathrm{sp}}+\mathcal L_{\mathrm{soot}}.
    \label{eq:training_objective}
\end{equation}
Image losses are averaged over cameras; camera parameters, spectral responses, calibration functions, and spatial weights remain fixed.

\noindent\textbf{Intensity fitting.}
For valid channel observations $\mathcal A_k$ in camera $k$, the RGB objective is
\begin{equation}
\begin{aligned}
    \mathcal L_{\mathrm{RGB}}
      &=\frac{1}{N_c}\sum_{k=1}^{N_c}\ell_{\mathrm{RGB}}^{(k)},\\
    \ell_{\mathrm{RGB}}^{(k)}
      &=\frac{1}{|\mathcal A_k|}\sum_{(p,c)\in\mathcal A_k}
      \left(\frac{\widehat I_{k,p,c}-I_{k,p,c}}{D}\right)^2,
\end{aligned}
    \label{eq:rgb_loss}
\end{equation}
where $D=4095$ is the fixed 12-bit DN scale. This normalization preserves relative intensity across cameras and channels. Invalid sensor values and annotated apparatus pixels are excluded from image supervision.

\noindent\textbf{Chromaticity consistency.}
Channel proportions provide complementary supervision to intensity fitting. With dark-corrected RGB vectors $\mathbf J=\mathbf I-\mathbf d$ and $\widehat{\mathbf J}=\widehat{\mathbf I}-\mathbf d$, define
\begin{equation}
    \mathbf c=\frac{\mathbf J}{\sum_c J_c},\qquad
    \widehat{\mathbf c}=
      \frac{\widehat{\mathbf J}}
      {\max(\sum_c\widehat J_c,\epsilon)}.
    \label{eq:chromaticity}
\end{equation}
The loss averages the squared channel error over each camera's reliable set $\mathcal P_{\mathrm{rel}}^{(k)}$:
\begin{equation}
\begin{aligned}
    \mathcal L_{\mathrm{ch}}
      &=\frac{1}{N_c}\sum_{k=1}^{N_c}\ell_{\mathrm{ch}}^{(k)},\\
    \ell_{\mathrm{ch}}^{(k)}
      &=\frac{\sum_{p\in\mathcal P_{\mathrm{rel}}^{(k)}}
      \|\widehat{\mathbf c}_{k,p}-\mathbf c_{k,p}\|_2^2}
      {3|\mathcal P_{\mathrm{rel}}^{(k)}|}.
\end{aligned}
    \label{eq:chromaticity_loss}
\end{equation}
Reliable pixels have sufficient observed signal in every channel and avoid saturation and invalid values; $\epsilon=1$ DN. Minibatch estimates use each camera's fixed reliable-pixel fraction to preserve the conditional mean.

\begin{figure}[htbp]
    \centering
    \includegraphics[width=0.95\columnwidth]{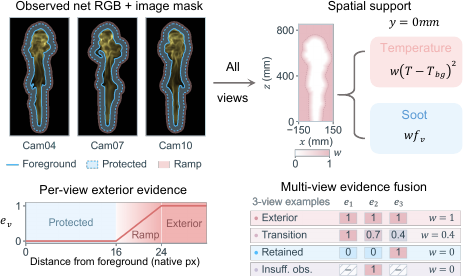}
    \caption{\textbf{Observation-derived spatial support.} Dark-corrected observations and masks define fixed multi-view exterior weights via protected margins and transition ramps. Invalid views abstain; insufficiently observed regions receive zero exterior weight.}
    \label{fig:spatial_support}
\end{figure}

\noindent\textbf{Observation-derived spatial support.}
\label{sec:spatial_support}
As illustrated in \cref{fig:spatial_support}, foreground masks are constructed from dark-corrected observations and projected into $\Omega$ using calibrated cameras. A location receives exterior evidence only when supported by at least three valid background views. Unobserved or invalid image regions abstain from this decision. Conservative margins and transition regions yield fixed weights $\omega_T(\mathbf x),\omega_f(\mathbf x)\in[0,1]$. These weights define
\begin{equation}
    \mathcal L_{\mathrm{sp}}=
      \lambda_T\langle\omega_T\overline T^{\,2}\rangle_\Omega
      +\lambda_{\mathrm{sp}}\langle\omega_f\overline f_v\rangle_\Omega,
    \label{eq:spatial_loss}
\end{equation}
where $\overline T=(T-T_{\mathrm{bg}})/T_{\mathrm{scale}}$, $\overline f_v=f_v/f_{\mathrm{ref}}$, and $\langle\cdot\rangle_\Omega$ denotes a uniform volume average over $\Omega$. Fully excluded exterior regions can additionally be assigned zero soot, using the same effective field for rendering and regularization; uncertain regions are retained. The temperature term provides background completion where soot emission offers little thermal information, rather than a direct temperature measurement.

\noindent\textbf{Soot regularization.}
To discourage excess soot and concentrated peaks, the objective includes
\begin{equation}
\begin{aligned}
    \mathcal L_{\mathrm{soot}}
      &=\lambda_m\langle\overline f_v\rangle_\Omega\\
      &\quad+\lambda_{\mathrm{cap}}
      \left\langle\left[\frac{f_v}{f_{\mathrm{cap}}}-1\right]_+^2
      \right\rangle_\Omega,
\end{aligned}
\label{eq:soot_loss}
\end{equation}
where $[z]_+=\max(z,0)$. The terms penalize total soot volume and softly discourage concentrations above $f_{\mathrm{cap}}$. Volume sampling is independent of image brightness; support-restricted mean-soot estimates are weighted by the fixed support's volume fraction to preserve full-domain normalization.

Supplementary Algorithm~S1 summarizes training.

\section{Experiments}
\setlength{\textfloatsep}{12pt}
\setlength{\dbltextfloatsep}{12pt}
\setlength{\floatsep}{10pt}
\setlength{\dblfloatsep}{10pt}
\setlength{\intextsep}{12pt}
\label{sec:experiments}

\subsection{Experimental Setup}
\label{sec:experimental_setup}

\begin{figure}[htbp]
    \centering
    \includegraphics[width=0.85\columnwidth]{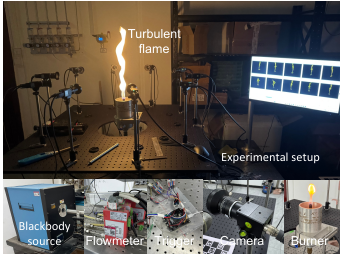}
    \caption{\textbf{Experimental acquisition platform.} Multi-camera acquisition of a propane diffusion flame and the associated calibration equipment.}
    \label{fig:experimental_platform}
\end{figure}

\noindent\textbf{Datasets.}
The main paper presents a synthetic methane-flame snapshot and a measured propane-flame snapshot, each with nine input views and Cam08 held out. Synthetic images are generated from CFD reference fields using the spectral imaging model, providing paired temperature and soot volume fraction references. The current synthetic snapshot at $t=0.51$ s serves as a development case. The measured dataset lacks independent three-dimensional field measurements.

\noindent\textbf{Experimental platform.}
\Cref{fig:experimental_platform} shows a propane diffusion burner without externally supplied coflow, operated at $1.147~\mathrm{L\,min^{-1}}$. Ten calibrated cameras form an approximately horizontal ring with roughly $36^{\circ}$ spacing, capturing images at a 3 ms exposure. Calibrated geometry defines the observation rays, while exposure-matched dark references and per-camera blackbody calibration map recorded signals to the spectral image-formation model. Supplementary Sec.~B provides further acquisition details.

\noindent\textbf{Baselines.}
RGB appearance baselines comprise Instant-NGP~\cite{mueller2022instant}, the vector--matrix variant of TensoRF~\cite{chen2022tensorf}, and 3DGS~\cite{kerbl2023gaussian}. These methods do not explicitly estimate temperature or soot. We independently implement nonnegative SIRT~\cite{gilbert1972iterative} and ASD-POCS~\cite{sidky2008image} based on their published algorithms. Our two-stage pipeline reconstructs RGB volume emission under an optically thin, emission-only model, followed by regularized pyrometry for temperature and soot volume fraction. These serve as classical tomography comparisons under simplified image formation. For a controlled representation comparison, our in-house Voxel baseline replaces only ThermoSight's neural field with a voxel grid, retaining its spectral emission--absorption renderer, camera responses and calibration, and full objective. Voxel+TV additionally includes total-variation (TV) regularization.

\noindent\textbf{Evaluation metrics.}
RGB metrics pool valid pixels and channels of held-out Cam08 on common rectified grids in calibrated linear RGB space. Physical-field methods are evaluated through RGB reprojections; display processing and background compositing are excluded from metric computation. Relative $\ell_2$ error and PSNR use fixed masks over the full valid domain and an observation-defined bright region (maximum channel signal $\geq10$ DN equivalent), with a shared observation-derived PSNR peak. Unless specified otherwise, reconstruction uses nine input views. Synthetic physical-field RMSE (K for temperature, ppm for soot) and relative $\ell_2$ errors use a common voxel-center grid and fixed reference masks for the full valid volume and a hot, sooty region ($T\geq1000$ K, $f_v\geq0.01$ ppm). Synthetic references quantify temperature and soot reconstruction accuracy; measured data assess practical applicability through RGB prediction on Cam08 and reconstruction across multiple propane flow rates.

\noindent\textbf{Implementation details.}
ThermoSight uses a Taichi hash encoder and an MLP with two 64-unit hidden layers, optimized from random initialization for 30,000 Adam updates on one NVIDIA RTX 5090. Appearance and voxel baselines share the 30,000-iteration budget, though per-iteration computation differs. Supplementary Sec.~B details acquisition, implementations, and evaluation; Secs.~C--D report convergence (Fig.~S4) and model-mismatch sensitivity (Fig.~S5).

\subsection{Evaluation on Synthetic Data}
\label{sec:simulated_evaluation}

\begin{figure}[htbp]
    \centering
    \includegraphics[width=\columnwidth]{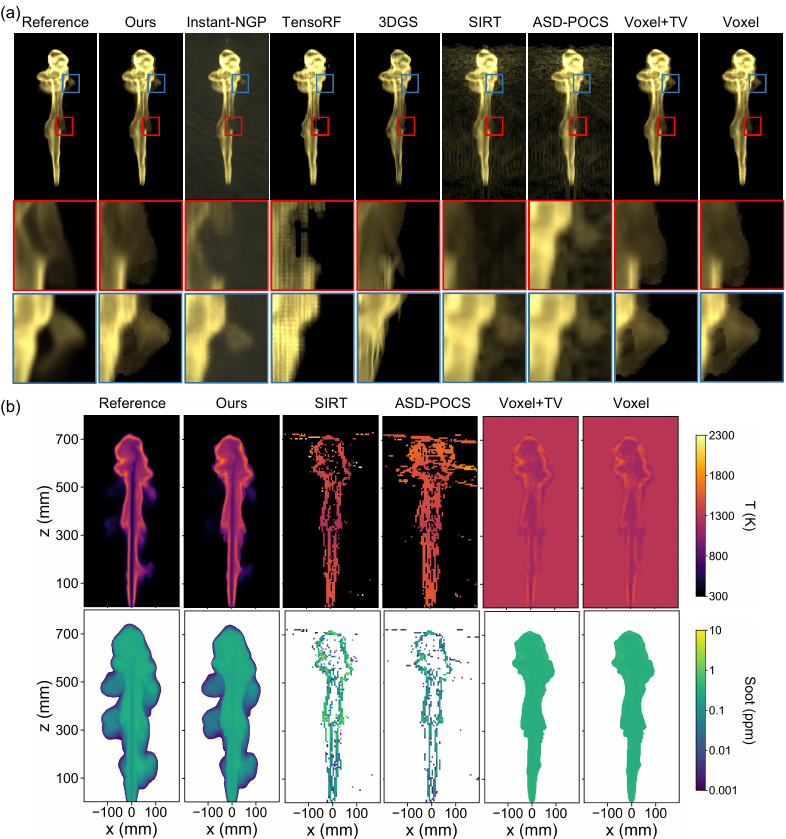}
    \caption{\textbf{Reconstruction on synthetic data.} (a) RGB and detail crops. (b) Temperature (top) and soot (bottom) at the labeled $y=0$ in Fig.~S1(a). SIRT and ASD-POCS use optically thin tomography followed by regularized pyrometry. Soot uses a logarithmic scale (0.001--10 ppm); white denotes below-range values or invalid inversions, and magenta denotes values above 10 ppm.}
    \label{fig:simulated_comparison}
\end{figure}

\setlength{\parskip}{0pt plus 2pt}
\noindent\textbf{Image Reconstruction.}
ThermoSight approaches the strongest appearance baseline in RGB fidelity: its full-domain PSNR is 0.10 dB below 3DGS, while its emitting-region PSNR is 0.23 dB higher (\cref{tab:simulated_results}). In \cref{fig:simulated_comparison}(a), ThermoSight preserves weak structures in the boxed crops. Instant-NGP retains colored background haze, TensoRF exhibits grid-like discontinuities, and 3DGS misses some boundary detail. Although these appearance-based reconstruction methods reproduce flame appearance well, they do not reveal the underlying temperature and soot fields.

SIRT/ASD-POCS show pronounced background artifacts, consistent with diffuse emission along poorly constrained rays and mismatch from omitted attenuation. Voxel and Voxel+TV retain a pale peripheral layer, suggesting residual boundary emission permitted by soft spatial support and compensating temperature--soot solutions. TV promotes smoothness but does not guarantee removal of boundary emission.

\begin{table*}[!tbp]
    \caption{\textbf{Quantitative reconstruction on synthetic data.} $e$ denotes relative $\ell_2$ error (\%); PSNR is in dB, temperature ($T$) RMSE in K, and soot ($f_v$) RMSE in ppm. $\dagger$ denotes tomography followed by regularized pyrometry; -- indicates no direct physical-field output. Column-wise best and second-best displayed values are \textbf{bold} and \underline{underlined}; ties share the same marker. 3DGS PSNR is derived from $e$ under the shared normalization.}
    \label{tab:simulated_results}
    \centering
    \begingroup
    \fontsize{8}{9}\selectfont
    \setlength{\aboverulesep}{0.25ex}
    \setlength{\belowrulesep}{0.3ex}
    \setlength{\tabcolsep}{1.2pt}
    \renewcommand{\arraystretch}{1}
    \begin{tabular*}{\textwidth}{@{\extracolsep{\fill}}C{66pt}*{4}{C{24pt}}C{35pt}C{30pt}C{37pt}C{38pt}C{35pt}C{30pt}C{37pt}C{38pt}@{}}
        \toprule
        \multirow{3}{=}[-4.0pt]{\centering Method} & \multicolumn{4}{c}{RGB reconstruction}
        & \multicolumn{8}{c}{Physical-field reconstruction} \\
        \cmidrule(lr){2-5}\cmidrule(lr){6-13}
        & \multicolumn{2}{c}{Full domain} & \multicolumn{2}{c}{Emitting}
        & \multicolumn{4}{c}{Full valid volume} & \multicolumn{4}{c}{Hot, sooty region} \\
        \cmidrule(lr){2-3}\cmidrule(lr){4-5}\cmidrule(lr){6-9}\cmidrule(lr){10-13}
        & $e\downarrow$ & PSNR$\uparrow$ & $e\downarrow$ & PSNR$\uparrow$
        & RMSE$_T\downarrow$ & $e_T\downarrow$ & RMSE$_{f_v}\downarrow$ & $e_{f_v}\downarrow$
        & RMSE$_T\downarrow$ & $e_T\downarrow$ & RMSE$_{f_v}\downarrow$ & $e_{f_v}\downarrow$ \\
        \midrule
        Instant-NGP~\cite{mueller2022instant}
        & 15.290 & 52.50 & 4.998 & 57.03
        & \multicolumn{4}{c}{--} & \multicolumn{4}{c}{--} \\
        TensoRF~\cite{chen2022tensorf}
        & 3.770 & 64.66 & 4.030 & 58.90
        & \multicolumn{4}{c}{--} & \multicolumn{4}{c}{--} \\
        3DGS~\cite{kerbl2023gaussian}
        & \textbf{3.485} & \textbf{65.34} & \underline{3.390} & \underline{60.41}
        & \multicolumn{4}{c}{--} & \multicolumn{4}{c}{--} \\
        SIRT$^\dagger$~\cite{gilbert1972iterative}
        & 10.502 & 55.76 & 8.570 & 52.35
        & \underline{192.953} & \underline{56.7027} & 3.9313 & 11061.9735
        & 569.327 & 44.6342 & 7.7141 & 2896.7055 \\
        ASD-POCS$^\dagger$~\cite{sidky2008image}
        & 10.671 & 55.62 & 8.514 & 52.41
        & 319.032 & 93.7533 & 2.1722 & 6112.1814
        & 487.646 & 38.2305 & 0.7561 & 283.9215 \\
        \rowcolor[HTML]{FCE8EC}
        Voxel+TV
        & 11.721 & 54.80 & 11.090 & 50.11
        & 984.950 & 289.4454 & \underline{0.02561} & \underline{72.0619}
        & \underline{228.054} & \underline{17.8790} & \underline{0.13042} & \underline{48.9737} \\
        \rowcolor[HTML]{FCE8EC}
        Voxel
        & 11.720 & 54.80 & 11.089 & 50.11
        & 984.951 & 289.4457 & \underline{0.02561} & \underline{72.0619}
        & \underline{228.054} & \underline{17.8790} & \underline{0.13042} & \underline{48.9737} \\
        \rowcolor[HTML]{E3EFFA}
        \textbf{Ours}
        & \underline{3.525} & \underline{65.24} & \textbf{3.300} & \textbf{60.64}
        & \textbf{29.377} & \textbf{8.6330} & \textbf{0.0106} & \textbf{29.8265}
        & \textbf{111.948} & \textbf{8.7765} & \textbf{0.0380} & \textbf{14.2693} \\
        \bottomrule
    \end{tabular*}
    \endgroup
\end{table*}

\noindent\textbf{Physical-Field Reconstruction.}
Optically thin SIRT/ASD-POCS neglect self-absorption and yield fragmented fields with large errors (\cref{fig:simulated_comparison}(b)). Channel-wise inversion errors may also propagate through pyrometry. With the same renderer and objective, Voxel improves soot and hot-region temperature recovery over SIRT/ASD-POCS, but has larger full-volume temperature errors (\cref{tab:simulated_results}). RGB agreement therefore does not reliably rank physical-field recovery. Supplementary Sec.~D shows that model mismatch can preserve RGB agreement despite physical-field bias.

Under the shared soft 300~K background prior, Voxel and Voxel+TV retain elevated background temperatures and nearly uniform interior soot fields. Their almost identical errors indicate little benefit from the tested TV setting. These residual deviations suggest that ambiguity persists despite the shared constraints. ThermoSight achieves the lowest errors in all physical-field metrics, reducing hot-region temperature and soot RMSE by 50.9\% and 70.9\%, respectively, relative to Voxel. This comparison primarily probes representation and optimization behavior; individual loss effects are assessed separately in \cref{tab:ablation_results}.

The prior penalizes temperature deviations without fixing background values. ThermoSight's low-temperature exterior is consistent with the broader spatial coupling induced by shared neural parameters. TV penalizes spatial differences rather than absolute temperature, so a uniformly elevated background can incur little TV penalty.

Voxel-based inversion discretizes physical fields on a prescribed grid, while our continuous neural representation supports direct coordinate queries and flexible sampling densities without retraining.

\setlength{\parskip}{0pt}
\subsection{Reconstruction from Measured Data}
\label{sec:experimental_reconstruction}

\begin{table}[!htbp]
    \caption{\textbf{RGB reconstruction on measured data.} Errors are relative $\ell_2$ percentages.}
    \label{tab:experimental_rgb}
    \centering
    \begingroup
    \fontsize{8}{9}\selectfont
    \setlength{\aboverulesep}{0.25ex}
    \setlength{\belowrulesep}{0.3ex}
    \setlength{\tabcolsep}{1.5pt}
    \renewcommand{\arraystretch}{1}
    \begin{tabular*}{0.94\columnwidth}{@{\extracolsep{\fill}}C{64pt}C{32pt}C{40pt}C{32pt}C{40pt}@{}}
        \toprule
        \multirow{2}{=}[-1.8pt]{\centering Method} & \multicolumn{2}{c}{Full valid domain}
        & \multicolumn{2}{c}{Emitting region} \\
        \cmidrule(lr){2-3}\cmidrule(lr){4-5}
        & Err. (\%)$\downarrow$ & PSNR (dB)$\uparrow$
        & Err. (\%)$\downarrow$ & PSNR (dB)$\uparrow$ \\
        \midrule
        Instant-NGP~\cite{mueller2022instant}
        & 8.9573 & 44.55 & 3.9659 & 46.85 \\
        TensoRF~\cite{chen2022tensorf}
        & 1.6731 & 59.13 & 1.6160 & 54.65 \\
        3DGS~\cite{kerbl2023gaussian}
        & \textbf{1.0111} & \textbf{63.50}
        & \underline{1.4572} & \underline{55.54} \\
        SIRT$^\dagger$~\cite{gilbert1972iterative}
        & 7.1059 & 46.56 & 6.1218 & 43.08 \\
        ASD-POCS$^\dagger$~\cite{sidky2008image}
        & 7.4647 & 46.14 & 6.7531 & 42.22 \\
        \rowcolor[HTML]{FCE8EC}
        Voxel
        & 11.3804 & 42.47 & 10.3442 & 38.52 \\
        \rowcolor[HTML]{FCE8EC}
        Voxel+TV
        & 11.3805 & 42.47 & 10.3446 & 38.52 \\
        \rowcolor[HTML]{E3EFFA}
        \textbf{Ours}
        & \underline{1.5078} & \underline{60.03}
        & \textbf{1.4133} & \textbf{55.81} \\
        \bottomrule
    \end{tabular*}
    \endgroup
\end{table}

\begin{figure}[t]
    \centering
    \includegraphics[width=\columnwidth]{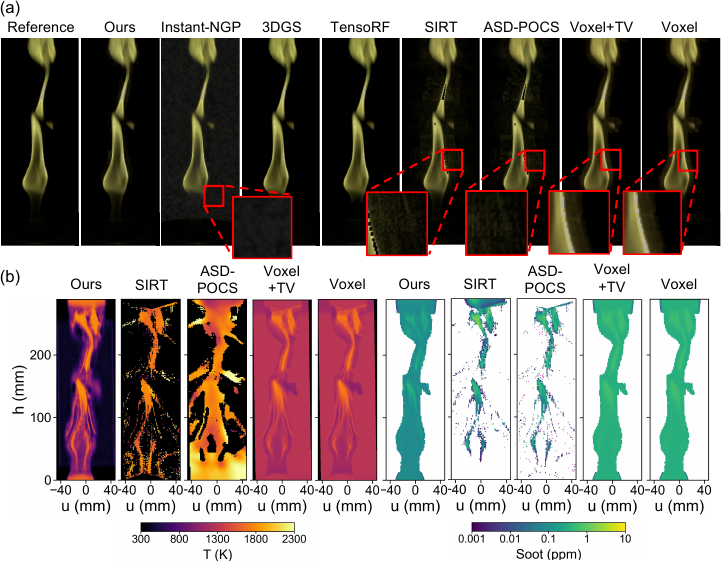}
    \caption{\textbf{Reconstruction from measured data.} (a) RGB. (b) Temperature (left) and soot (right) at $v=0$ in Fig.~S1(b). Physical-field baselines and soot display conventions follow \cref{fig:simulated_comparison}(b).}
    \label{fig:experimental_comparison}
\end{figure}

\noindent\textbf{Image Reconstruction.}
In \cref{fig:experimental_comparison}(a), Instant-NGP retains haze, SIRT/ASD-POCS show background artifacts, and Voxel and Voxel+TV retain a pale peripheral layer. ThermoSight preserves the measured flame structures with a dark background. 3DGS leads in the full-domain metrics, while ThermoSight leads in the emitting region (\cref{tab:experimental_rgb}).

\noindent\textbf{Physical-Field Reconstruction.}
In \cref{fig:experimental_comparison}(b), SIRT/ASD-POCS again produce fragmented fields, while Voxel and Voxel+TV retain elevated background temperatures and nearly uniform interior soot fields. ThermoSight recovers coherent temperature and soot fields while closely reproducing RGB observations from Cam08. Across the RGB and physical-field comparisons, these reconstruction patterns are consistent with those observed on synthetic data. Together with RGB agreement on Cam08, this consistency supports the calibrated pipeline's applicability to measured data.

Supplementary Sec.~A provides quantitative evaluations and field visualizations for additional synthetic snapshots and for measurements at different propane flow rates.

\subsection{Sparse-View Reconstruction}
\label{sec:sparse_view}

\begin{table*}[tbp]
    \caption{\textbf{Sparse-view reconstruction on synthetic data.}  RGB uses held-out Cam08; PSNR is derived from $e$ using nine-view normalization. Annotations: RMSE increases (\%) / PSNR drops (dB) vs. nine views.}
    \label{tab:sparse_results}
    \centering
    \begingroup
    \fontsize{8}{9}\selectfont
    \newcommand{\sparsedelta}[2]{#1\raisebox{-1pt}{\fontsize{4.5}{4.5}\selectfont\color{red!70!black}$\uparrow#2$}}
    \newcommand{\sparseloss}[2]{#1\raisebox{-1pt}{\fontsize{4.5}{4.5}\selectfont\color{blue!70!black}$\downarrow#2$}}
    \setlength{\aboverulesep}{0.3ex}
    \setlength{\belowrulesep}{0.4ex}
    \setlength{\tabcolsep}{0.3pt}
    \renewcommand{\arraystretch}{1.05}
    \begin{tabular*}{\textwidth}{@{\extracolsep{\fill}}C{24pt}C{44pt}C{40pt}C{44pt}C{40pt}C{50pt}C{26pt}C{43pt}C{26pt}C{47pt}C{26pt}C{43pt}C{26pt}@{}}
        \toprule
        \multirow{3}{=}[-4.0pt]{\centering Views} & \multicolumn{4}{c}{RGB reconstruction}
        & \multicolumn{8}{c}{Physical-field reconstruction} \\
        \cmidrule(lr){2-5}\cmidrule(lr){6-13}
        & \multicolumn{2}{c}{Full region}
        & \multicolumn{2}{c}{Emitting region}
        & \multicolumn{4}{c}{Full valid volume}
        & \multicolumn{4}{c}{Hot, sooty region} \\
        \cmidrule(lr){2-3}\cmidrule(lr){4-5}
        \cmidrule(lr){6-9}\cmidrule(lr){10-13}
        & $e\downarrow$ & PSNR$\uparrow$
        & $e\downarrow$ & PSNR$\uparrow$
        & RMSE$_T\downarrow$ & $e_T\downarrow$ & RMSE$_{f_v}\downarrow$ & $e_{f_v}\downarrow$
        & RMSE$_T\downarrow$ & $e_T\downarrow$ & RMSE$_{f_v}\downarrow$ & $e_{f_v}\downarrow$ \\
        \midrule
        \rowcolor{black!7}

        9 & \textbf{3.5250} & \textbf{65.24} & \textbf{3.3000} & \textbf{60.64}
          & \textbf{29.377} & \textbf{8.6330} & \textbf{0.0106} & \textbf{29.8265}
          & \textbf{111.948} & \textbf{8.7765} & \textbf{0.0380} & \textbf{14.2693} \\

        8 & \underline{5.1774} & \sparseloss{\underline{61.90}}{3.34} & \underline{4.8926} & \sparseloss{\underline{57.22}}{3.42}
          & \sparsedelta{\underline{36.858}}{25.5} & \underline{10.8315} & \sparsedelta{\underline{0.0114}}{7.5} & \underline{32.1171}
          & \sparsedelta{\underline{112.374}}{0.4} & \underline{8.8099} & \sparsedelta{\underline{0.0654}}{72.1} & \underline{24.5513} \\

        6 & 5.3582 & \sparseloss{61.60}{3.64} & 5.0521 & \sparseloss{56.94}{3.70}
          & \sparsedelta{54.027}{83.9} & 15.8770 & \sparsedelta{0.0131}{23.6} & 36.8373
          & \sparsedelta{119.677}{6.9} & 9.3824 & \sparsedelta{0.0879}{131.3} & 32.9912 \\

        4 & 7.0516 & \sparseloss{59.22}{6.02} & 6.6966 & \sparseloss{54.49}{6.15}
          & \sparsedelta{150.247}{411.4} & 44.1530 & \sparsedelta{0.0143}{34.9} & 40.1523
          & \sparsedelta{128.573}{14.9} & 10.0799 & \sparsedelta{0.1001}{163.4} & 37.6013 \\
        \bottomrule
    \end{tabular*}
    \endgroup
\end{table*}

\begin{figure}[t]
    \centering
    \includegraphics[width=0.9\columnwidth]{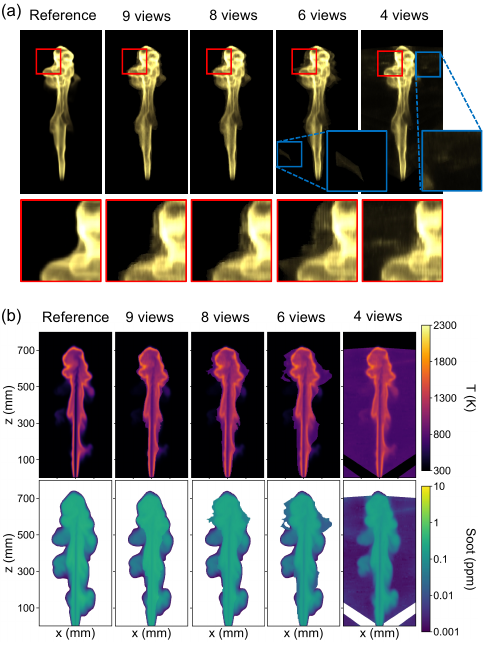}
    \caption{\textbf{Synthetic sparse-view reconstruction.} (a) RGB with crops. (b) Temperature (top) and soot (bottom), following \cref{fig:simulated_comparison}(b) for the slice plane and soot scale.}
    \label{fig:sparse_view_reconstruction}
\end{figure}

With Cam08 held out throughout, fewer input views broaden peripheral RGB emission and distort physical-field boundaries (\cref{fig:sparse_view_reconstruction}). Detached RGB artifacts appear at six and four views, while four views produce broad exterior field leakage. The increasing full-domain errors in \cref{tab:sparse_results} are consistent with weaker angular localization and exterior support, allowing spurious temperature--soot distributions to fit similar ray-integrated signals.

Hot-region temperature RMSE also rises to 119.677 K at six views and 128.573 K at four views. Thus, sparse observations degrade both interior temperature recovery and exterior field localization. For the tested scene and camera layout, at least six views with broad angular coverage are recommended for more reliable physical-field reconstruction. These results suggest potential for physical-field reconstruction in aero-engine combustors, gas turbines, industrial furnaces, fire scenarios, and other settings with limited camera access.

\subsection{Ablation Studies}
\label{sec:ablation_studies}

\begin{table*}[t]
    \caption{\textbf{Recorded ablation results on synthetic data.} RMSE annotations show increases (\%) over Full model.}
    \label{tab:ablation_results}
    \centering
    \begingroup
    \fontsize{8}{9}\selectfont
    \newcommand{\ablationdelta}[2]{#1\raisebox{-1pt}{\fontsize{5}{5}\selectfont\color{red!70!black}$\uparrow#2$}}
    \setlength{\aboverulesep}{0.3ex}
    \setlength{\belowrulesep}{0.4ex}
    \setlength{\tabcolsep}{0.8pt}
    \renewcommand{\arraystretch}{1.05}
    \begin{tabular*}{\textwidth}{@{\extracolsep{\fill}}C{71pt}*{4}{C{24pt}}C{52pt}C{30pt}C{44pt}C{28pt}C{48pt}C{28pt}C{44pt}C{28pt}@{}}
        \toprule
        \multirow{3}{=}[-4.0pt]{\centering Configuration} & \multicolumn{4}{c}{RGB reconstruction}
        & \multicolumn{8}{c}{Stored physical-field versions} \\
        \cmidrule(lr){2-5}\cmidrule(lr){6-13}
        & \multicolumn{2}{c}{Full image}
        & \multicolumn{2}{c}{Emitting region}
        & \multicolumn{4}{c}{Full valid volume}
        & \multicolumn{4}{c}{Hot, sooty region} \\
        \cmidrule(lr){2-3}\cmidrule(lr){4-5}
        \cmidrule(lr){6-9}\cmidrule(lr){10-13}
        & $e\downarrow$ & PSNR$\uparrow$
        & $e\downarrow$ & PSNR$\uparrow$
        & RMSE$_T\downarrow$ & $e_T\downarrow$ & RMSE$_{f_v}\downarrow$ & $e_{f_v}\downarrow$
        & RMSE$_T\downarrow$ & $e_T\downarrow$ & RMSE$_{f_v}\downarrow$ & $e_{f_v}\downarrow$ \\
        \midrule
        \rowcolor[HTML]{E3EFFA}

        Full model & \textbf{3.525} & \textbf{65.24} & \textbf{3.300} & \textbf{60.64}
          & \textbf{29.377} & \textbf{8.6330} & \textbf{0.0106} & \textbf{29.8265}
          & \textbf{111.948} & \textbf{8.7765} & \textbf{0.0380} & \textbf{14.2693} \\

        w/o chromaticity loss & 3.560 & 65.15 & \underline{3.3114} & \underline{60.61}
          & \ablationdelta{\underline{40.937}}{39.4} & \underline{12.0300} & \ablationdelta{\underline{0.0140}}{32.1} & \underline{39.3034}
          & \ablationdelta{169.902}{51.8} & 13.3200 & \ablationdelta{\underline{0.0513}}{35.0} & \underline{19.2536} \\

        w/o mean-soot prior & 3.536 & 65.22 & 3.3498 & 60.51
          & \ablationdelta{49.682}{69.1} & 14.6000 & \ablationdelta{0.0235}{121.7} & 66.0463
          & \ablationdelta{209.317}{87.0} & 16.4100 & \ablationdelta{0.1102}{190.0} & 41.3791 \\

        w/o spatial support & 3.737 & 64.73 & 3.5319 & 60.05
          & \ablationdelta{639.264}{2076.1} & 187.8600 & \ablationdelta{0.0262}{147.2} & 73.7916
          & \ablationdelta{\underline{156.637}}{39.9} & \underline{12.2800} & \ablationdelta{0.1675}{340.8} & 62.8909 \\

        3-channel transport & \underline{3.527} & \underline{65.23} & 3.3382 & 60.54
          & \ablationdelta{56.760}{93.2} & 16.6800 & \ablationdelta{0.0185}{74.5} & 52.1890
          & \ablationdelta{234.445}{109.4} & 18.3800 & \ablationdelta{0.0719}{89.2} & 27.0106 \\
        \bottomrule
    \end{tabular*}
    \endgroup
\end{table*}

\begin{figure}[htbp]
    \centering
    \includegraphics[width=\columnwidth]{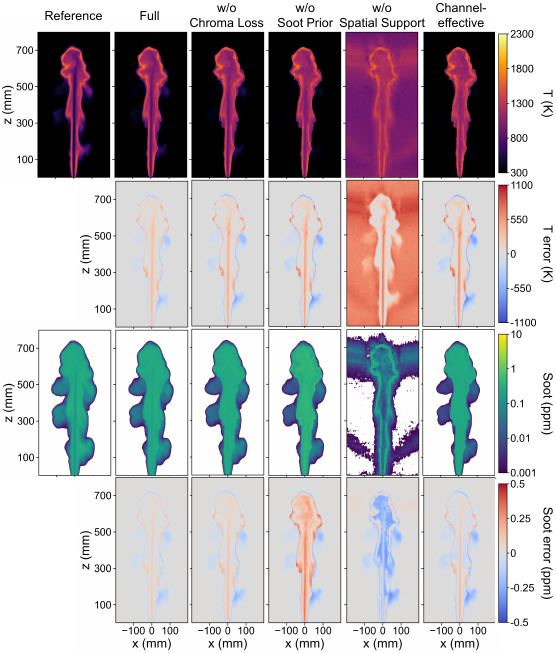}
    \caption{\textbf{Qualitative comparison of ablation configurations.} On the plane in \cref{fig:simulated_comparison}(b), rows show temperature, its signed error, soot, and its signed error for the stored visualization versions. Errors are prediction minus reference.}
    \label{fig:ablation_study}
\end{figure}

Despite similar full-image PSNR (64.73--65.24 dB), ablations yield different internal fields (\cref{tab:ablation_results,fig:ablation_study}). Removing chromaticity loss raises hot-region soot RMSE from 0.0380 to 0.0513 ppm and relaxes channel-proportion supervision, while the spectral renderer still couples RGB responses to shared fields.

Removing the mean-soot prior raises hot-region soot RMSE to 0.1102 ppm. Excess soot and lower temperatures suggest compensation for weaker thermal emission (\cref{eq:soot_coefficients}), which the prior discourages. Spatial support improves both background temperature completion and hot-region recovery. Removing it raises full-volume temperature RMSE from 29.377 to 639.264 K, consistent with exterior temperature drift and soot leakage. The hot-region increase from 111.948 to 156.637 K also indicates a benefit to internal temperature recovery.

The channel-effective variant retains response-weighted emission but uses three attenuation coefficients. Despite similar RGB fitting, hot-region temperature RMSE rises from 111.948 to 234.445 K. This approximation can alter the temperature--soot trade-off because attenuation and spectral integration do not generally commute (see Supplementary Sec.~B.3).

\section{Conclusion and Future Work}
\label{sec:conclusion}

We presented ThermoSight, a physics-guided neural framework for reconstructing continuous 3D temperature and soot volume fraction fields from limited calibrated flame images without field-level supervision. Calibrated spectral emission--absorption modeling and complementary constraints reduce temperature--soot ambiguity, extending appearance reconstruction toward scientific imaging of internal physical states. Synthetic evaluations show the lowest physical-field errors among the tested baselines with competitive RGB fidelity, highlighting that image agreement alone does not establish physical accuracy. Reconstructions from measured data yield coherent fields consistent with the observed images. The continuous representation enables arbitrary spatial queries and flexible sampling densities. Reconstruction remains sensitive to sparse observations and model mismatch. We will explore multispectral infrared imaging and gas-band radiative transfer to recover CO$_2$ and H$_2$O concentration fields, and temporal representations with flow constraints for 4D reconstruction.

\iftoggle{cvprfinal}{%
    \section*{Acknowledgements}
    This work was supported by the National Natural Science Foundation of China (No. 52576180).
}{}

\flushbottom
\section*{LLM Usage Statement}
OpenAI ChatGPT/GPT and Codex were used at the assistance level to edit and refine author-produced material related to manuscript writing, related work and citations, code and implementation, experimental design, execution and result analysis, figures and diagrams, and data preparation. OpenAI ChatGPT/GPT also provided reviewer-style feedback on the manuscript. The research ideas, hypotheses, scientific claims, and conclusions originated with the authors. The authors take full responsibility for all submitted content.

\clearpage
\setcounter{section}{0}
\setcounter{figure}{0}
\setcounter{table}{0}
\setcounter{equation}{0}
\setcounter{algorithm}{0}
\renewcommand{\thesection}{\Alph{section}}
\renewcommand{\thefigure}{S\arabic{figure}}
\renewcommand{\thetable}{S\arabic{table}}
\renewcommand{\theequation}{S\arabic{equation}}
\renewcommand{\thealgorithm}{S\arabic{algorithm}}
\renewcommand{\theHsection}{supp.\Alph{section}}
\renewcommand{\theHfigure}{supp.\arabic{figure}}
\renewcommand{\theHtable}{supp.\arabic{table}}
\renewcommand{\theHequation}{supp.\arabic{equation}}
\providecommand{\theHalgorithm}{\arabic{algorithm}}
\renewcommand{\theHalgorithm}{supp.\arabic{algorithm}}
\newcolumntype{C}[1]{>{\centering\arraybackslash}p{#1}}
\maketitlesupplementary
\flushbottom
\setlength{\parskip}{0pt plus 1pt}
\setlength{\textfloatsep}{20pt plus 2pt minus 4pt}
\setlength{\dbltextfloatsep}{20pt plus 2pt minus 4pt}
\setlength{\floatsep}{12pt plus 2pt minus 2pt}
\setlength{\dblfloatsep}{12pt plus 2pt minus 2pt}
\setlength{\intextsep}{12pt plus 2pt minus 2pt}
\input{sec/supp_contents}
\section{Reconstruction Across Operating Conditions}
\label{sec:supp_additional_conditions}

\begin{figure*}[tp]
    \centering
    \includegraphics[width=\textwidth,height=0.9\textheight,keepaspectratio]{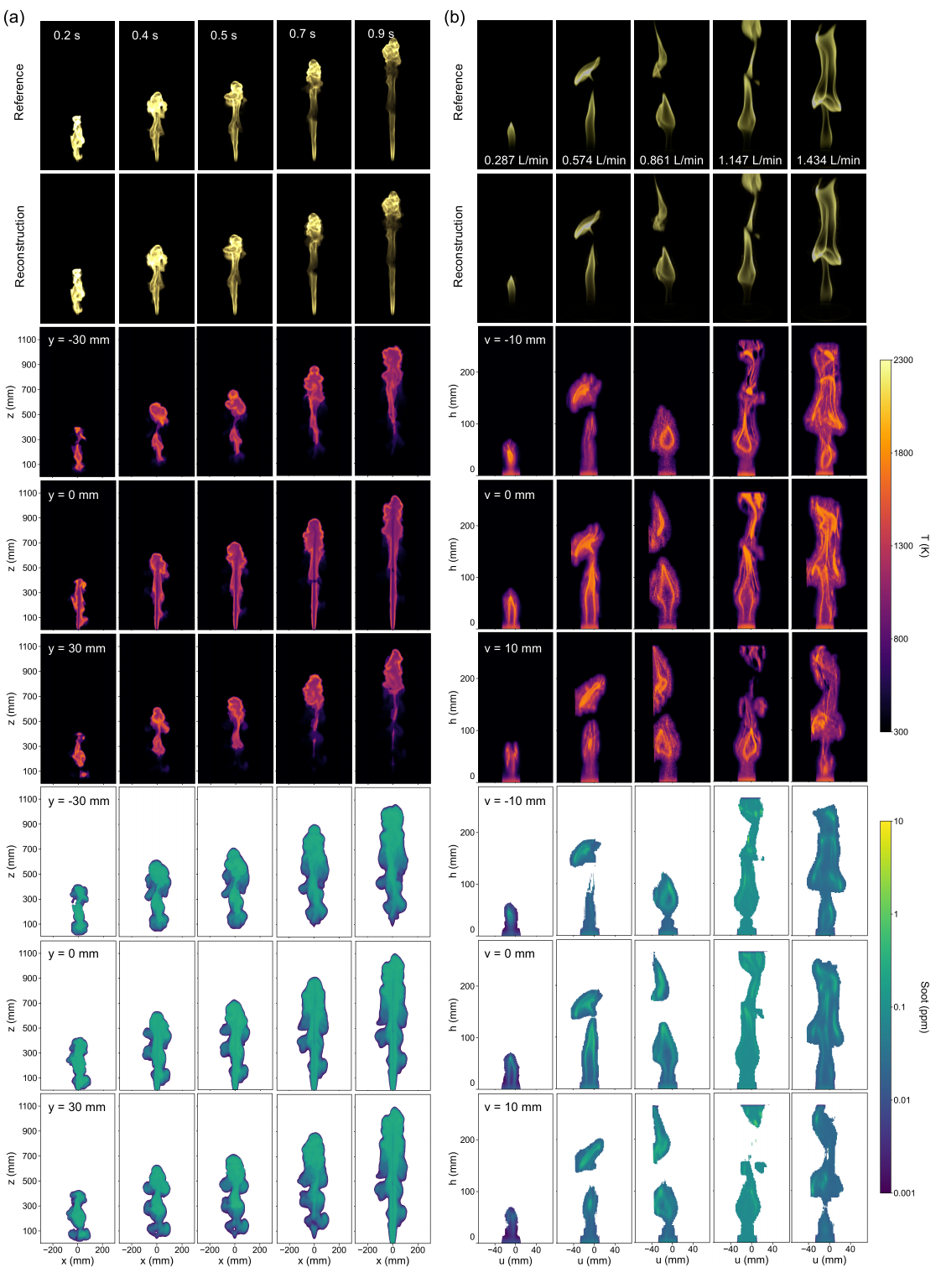}
    \captionsetup{skip=4pt}
    \caption{\textbf{Reconstruction from synthetic and measured flame data.} (a) Synthetic frames at $t=0.2,0.4,0.5,0.7,0.9$~s. (b) Experimental propane flames at 0.287, 0.574, 0.861, 1.147, and 1.434~L/min. The first two rows show reference and reconstructed RGB images, respectively. Rows 3--5 show temperature slices at three spatial offsets, and rows 6--8 show soot volume fraction slices at the same offsets. The zero offsets indicate the main-paper slice planes. In (a), the labels give offsets from the $x$--$z$ plane at world $y=-3.08$~mm. In (b), $v$ is the offset along the Cam04 optical axis from the plane through the reconstruction domain center. Temperature is in K; soot volume fraction is in ppm on a logarithmic scale. In (b), the RGB references are the held-out camera images.}
    \label{fig:supp_additional_conditions}
\end{figure*}

\begin{figure*}[tp]
    \centering
    \includegraphics[width=\textwidth,height=0.65\textheight,keepaspectratio]{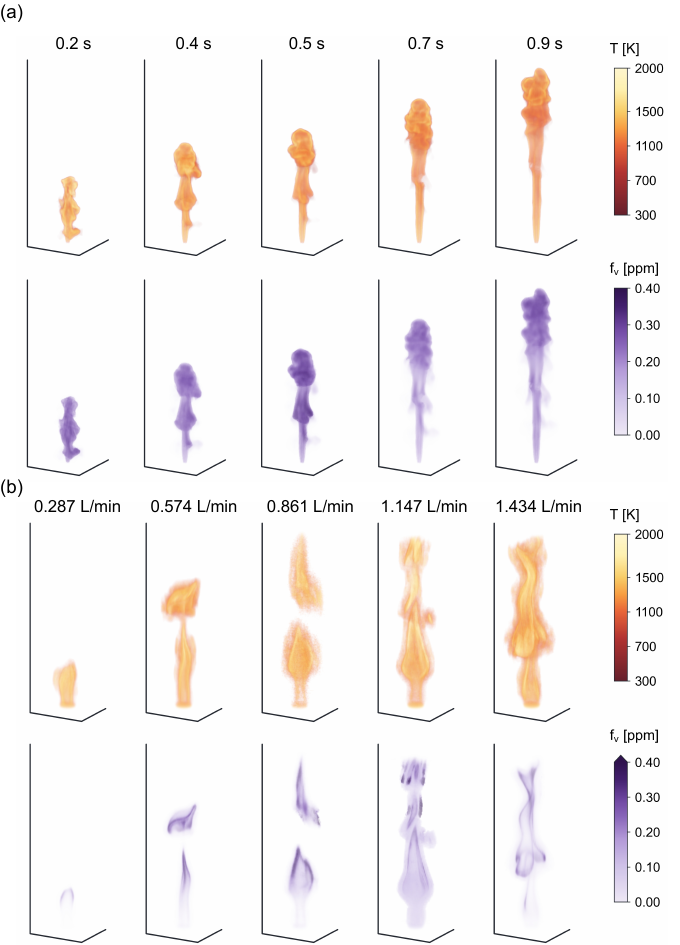}
    \captionsetup{skip=4pt}
    \caption{\textbf{Three-dimensional physical-field reconstructions across operating conditions.} (a) Synthetic frames at $t=0.2,0.4,0.5,0.7,0.9$~s. (b) Experimental propane flames at 0.287, 0.574, 0.861, 1.147, and 1.434~L/min. Each panel shows reconstructed temperature fields (top, K) and soot volume fraction fields (bottom, ppm), with the respective color scales on the right. Cases correspond to those in \cref{fig:supp_additional_conditions}.}
    \label{fig:supp_additional_3d_fields}

    \vspace{8pt}
    \includegraphics[width=\textwidth]{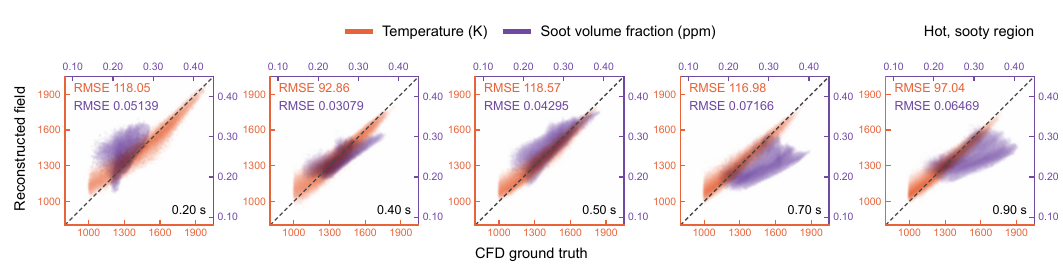}
    \captionsetup{skip=4pt}
    \caption{\textbf{Pointwise physical-field agreement across synthetic times.} Each panel compares reconstructed values with CFD references in the hot, sooty region at the indicated time. Orange points use the bottom/left temperature axes (K); purple points use the top/right soot-volume-fraction axes (ppm). Dashed diagonals indicate equality. The annotated RMSEs correspond to the hot-region values in \cref{tab:supp_simulated_results}.}
    \label{fig:supp_pointwise_fields}
\end{figure*}

\Cref{fig:supp_additional_conditions} extends the main-paper evaluation to multiple times in the synthetic sequence and to measurements at five propane flow rates. RGB comparisons and physical-field slices are shown for each case.
\Cref{fig:supp_additional_3d_fields} visualizes the corresponding three-dimensional temperature and soot fields, complementing the cross-sectional views.
\Cref{tab:supp_simulated_results} reports quantitative RGB and physical-field errors for the five synthetic snapshots.
\Cref{tab:supp_experimental_rgb} reports RGB errors on measured data across the five flow rates.
\Cref{fig:supp_pointwise_fields} compares reconstructed and CFD reference values pointwise in the hot, sooty region at all five synthetic times, using the corresponding evaluation masks and RMSEs in \cref{tab:supp_simulated_results}.

Across the synthetic sequence, ThermoSight reproduces the evolving flame appearance and reconstructs spatially coherent temperature and soot distributions. Reconstructions across propane flow rates also capture changes in flame shape and internal field structure. These results extend the qualitative evaluation beyond the snapshots presented in the main paper. The temperature point clouds generally follow the identity line, while soot exhibits broader off-diagonal distributions, particularly at 0.7 and 0.9 s. These plots complement the spatial visualizations by revealing pointwise deviations.

\begin{table*}[tp]
    \caption{\textbf{Quantitative reconstruction across synthetic snapshots.} $e$ denotes relative $\ell_2$ error (\%); PSNR is in dB, temperature ($T$) RMSE in K, and soot ($f_v$) RMSE in ppm. Reconstruction uses nine training views, with RGB metrics evaluated on held-out Cam08; the emitting region comprises valid pixels with an observed maximum channel signal $\geq 10$ DN.}
    \label{tab:supp_simulated_results}
    \centering
    \begingroup
    \fontsize{8}{9.6}\selectfont
    \setlength{\aboverulesep}{0.3ex}
    \setlength{\belowrulesep}{0.4ex}
    \setlength{\tabcolsep}{1.2pt}
    \renewcommand{\arraystretch}{1.0}
    \begin{tabular*}{\textwidth}{@{\extracolsep{\fill}}C{32pt}*{4}{C{24pt}}C{35pt}C{30pt}C{37pt}C{30pt}C{35pt}C{30pt}C{37pt}C{30pt}@{}}
        \toprule
        \multirow{3}{=}[-4.0pt]{\centering Time (s)} & \multicolumn{4}{c}{RGB reconstruction}
        & \multicolumn{8}{c}{Physical-field reconstruction} \\
        \cmidrule(lr){2-5}\cmidrule(lr){6-13}
        & \multicolumn{2}{c}{Full domain} & \multicolumn{2}{c}{Emitting region}
        & \multicolumn{4}{c}{Full volume} & \multicolumn{4}{c}{Hot, sooty region} \\
        \cmidrule(lr){2-3}\cmidrule(lr){4-5}\cmidrule(lr){6-9}\cmidrule(lr){10-13}
        & $e\downarrow$ & PSNR$\uparrow$ & $e\downarrow$ & PSNR$\uparrow$
        & RMSE$_T\downarrow$ & $e_T\downarrow$ & RMSE$_{f_v}\downarrow$ & $e_{f_v}\downarrow$
        & RMSE$_T\downarrow$ & $e_T\downarrow$ & RMSE$_{f_v}\downarrow$ & $e_{f_v}\downarrow$ \\
        \midrule
        0.20 & 5.87 & 53.14 & 3.25 & 41.43
             & 19.94 & 6.27 & 0.00769 & 42.91
             & 118.05 & 8.28 & 0.05139 & 23.02 \\
        0.40 & 4.32 & 59.10 & 2.36 & 49.80
             & 25.88 & 7.70 & 0.00835 & 27.57
             & 92.86 & 6.91 & 0.03079 & 12.27 \\
        0.50 & 6.00 & 60.22 & 3.26 & 51.26
             & 29.13 & 8.58 & 0.01117 & 31.82
             & 118.57 & 9.19 & 0.04295 & 16.21 \\
        0.70 & 6.62 & 61.53 & 3.56 & 53.38
             & 33.25 & 9.58 & 0.01487 & 33.93
             & 116.98 & 9.34 & 0.07166 & 25.27 \\
        0.90 & 6.39 & 63.81 & 3.50 & 56.37
             & 35.52 & 10.05 & 0.01740 & 33.47
             & 97.04 & 7.88 & 0.06469 & 21.97 \\
        \bottomrule
    \end{tabular*}
    \endgroup
\end{table*}

\begin{table}[tp]
    \caption{\textbf{RGB reconstruction on measured data at five propane flow rates.} RGB metrics use held-out Cam08. Errors are relative $\ell_2$ percentages; PSNR is in dB. $\dagger$ denotes the same propane flow rate as in the main paper.}
    \label{tab:supp_experimental_rgb}
    \centering
    \begingroup
    \fontsize{8}{9.6}\selectfont
    \setlength{\aboverulesep}{0.3ex}
    \setlength{\belowrulesep}{0.4ex}
    \setlength{\tabcolsep}{1.5pt}
    \renewcommand{\arraystretch}{1.0}
    \begin{tabular*}{0.98\columnwidth}{@{\extracolsep{\fill}}C{52pt}C{34pt}C{42pt}C{34pt}C{42pt}@{}}
        \toprule
        \multirow{2}{=}[-1.8pt]{\centering Propane flow (L/min)}
        & \multicolumn{2}{c}{Full valid domain}
        & \multicolumn{2}{c}{Emitting region} \\
        \cmidrule(lr){2-3}\cmidrule(lr){4-5}
        & Err. (\%)$\downarrow$ & PSNR (dB)$\uparrow$
        & Err. (\%)$\downarrow$ & PSNR (dB)$\uparrow$ \\
        \midrule
        0.287 & 1.8015 & 67.23 & 1.7615 & 52.33 \\
        0.574 & 1.4001 & 60.72 & 1.3749 & 51.35 \\
        0.861 & 2.4259 & 57.13 & 2.3150 & 48.95 \\
        \rowcolor{black!7}
        1.147$^\dagger$ & 1.5078 & 60.03 & 1.4133 & 55.81 \\
        1.434 & 1.5333 & 57.21 & 1.4988 & 51.27 \\
        \bottomrule
    \end{tabular*}
    \endgroup
\end{table}

\section{Acquisition and Implementation}
\label{sec:supp_acquisition_and_implementation}
\subsection{Acquisition Details}
\label{sec:supp_acquisition}

A propane diffusion burner operates without externally supplied coflow. For the measured snapshot in the main paper, a Sevenstar CS230 controller sets the flow to $1.147~\mathrm{L\,min^{-1}}$. Images are captured after 15 s of stabilization. Ten calibrated cameras form a horizontal ring at radii of 0.40--0.42 m with approximately $36^{\circ}$ spacing; nine provide reconstruction inputs and Cam08 is held out. RAW images have $2448\times2048$ pixels, 12-bit depth, and 3 ms exposure. Exposure-matched dark references and per-camera blackbody calibration establish the radiometric mapping.

\noindent\textbf{Radiometric mapping.}
For each camera and channel, the calibration pairs a reference blackbody signal with the corresponding dark-corrected RAW DN. The model input $x$ is the exposure time times the integral of the reference spectral radiance against the effective channel response. A fixed quadratic $g_{k,c}(x)=a_{k,c}x^2+b_{k,c}x+q_{k,c}$ maps this input to dark-corrected DN, with the input scale absorbed into the coefficients. During reconstruction, the same mapping is applied to the rendered spectrum after channel integration and exposure scaling. The dark reference is then added back to predict RAW DN.

\subsection{Implementation Details}
\label{sec:supp_implementation}

\subsubsection{ThermoSight}
\label{sec:supp_thermosight_implementation}

ThermoSight uses PyTorch with a Taichi hash encoder (16 levels, two features per entry, resolutions 16--256, at most $2^{19}$ entries per level) and a shared MLP with two 64-unit SiLU hidden layers. Temperature is bounded to 300--2300 K, with soot output scale $5\times10^{-7}$. Loss normalization uses $T_{\mathrm{bg}}=300$ K, $T_{\mathrm{scale}}=1000$ K, and $f_{\mathrm{ref}}=10^{-6}$; the soft soot cap is $3\times10^{-6}$. Both datasets use the same architecture and objective. Training uses seed 42 (Algorithm~\ref{alg:supp_thermosight_training}).

\noindent\textbf{Runtime.} On the synthetic dataset, ThermoSight takes approximately 11 minutes to complete 30,000 Adam optimization steps on a single NVIDIA RTX 5090 GPU with 32~GB of memory.

\begin{algorithm}[t]
    \caption{ThermoSight training}
    \label{alg:supp_thermosight_training}
    \small
    \begin{algorithmic}[1]
        \Require Training RGB images, valid-pixel masks, dark references,
                 calibrated cameras and responses, domain $\Omega$
        \Ensure Queryable temperature--soot fields $\Phi_\theta=(T_\theta,f_{v,\theta})$
        \State Build fixed foreground masks and multi-view spatial support from training images only
        \State Initialize the hash encoder, MLP, and Adam; freeze cameras and calibration
        \For{$s=1,\ldots,30{,}000$}
            \State Set $\eta(s)$ and $\lambda_{\mathrm{ch}}(s)$ using the schedules below
            \State Cycle cameras every 16 updates; sample 256 valid rays
            \State Draw one stratified point in each of 128 segments per ray inside $\Omega$, plus 1,024 uniform box points
            \State Query $T_\theta,f_{v,\theta}$; apply fixed exterior soot exclusion
            \State Compute spectral emission $j_\lambda$ and absorption $\kappa_\lambda$
            \State Integrate emission--absorption with the analytic segment solution to obtain ray radiance $\widehat L_\lambda$
            \State Apply spectral responses, exposure, and radiometric calibration to obtain predicted RGB $\widehat{\mathbf I}_k$
            \State Compute $\mathcal L_{\mathrm{RGB}}$ on valid channels, scaled by $D=4095$
            \If{$s>12{,}000$}
                \State Compute $\mathcal L_{\mathrm{ch}}$ from reliable dark-corrected channel proportions
                \State Estimate mean soot from 1,024 separate ROI samples, with full-domain volume-fraction correction
            \Else
                \State Set $\mathcal L_{\mathrm{ch}}=0$; use box samples for mean soot
            \EndIf
            \State Compute $\mathcal L_{\mathrm{sp}}$ with fixed support weights and $\mathcal L_{\mathrm{soot}}$ with the mean and soft-cap terms
            \State $\mathcal L\gets\mathcal L_{\mathrm{RGB}}+
                \lambda_{\mathrm{ch}}(s)\mathcal L_{\mathrm{ch}}+
                \mathcal L_{\mathrm{sp}}+\mathcal L_{\mathrm{soot}}$
            \State Backpropagate $\nabla_\theta\mathcal L$; update $\theta$ with Adam at $\eta(s)$
        \EndFor
        \State \Return $\Phi_\theta$ for arbitrary queries in $\Omega$
    \end{algorithmic}
\end{algorithm}

Loss weights are $\lambda_T=10^{-5}$, $\lambda_{\mathrm{sp}}=3\times10^{-5}$, $\lambda_m=1.2\times10^{-4}$, and $\lambda_{\mathrm{cap}}=10^{-3}$. Chromaticity is disabled through update 12,000, then uses $\lambda_{\mathrm{ch}}\simeq8.44249\times10^{-3}$. The encoder and MLP share a learning rate of $10^{-3}$ through update 10,000. At update 10,001 it drops to $10^{-4}$, follows cosine annealing to $10^{-5}$ by update 12,000, and remains fixed thereafter.

\noindent\textbf{Rendering and sampling.}
Spectral integration uses 1 nm composite trapezoidal quadrature on a calibrated 350--1100 nm grid, followed by an ideal 650 nm short-pass filter. The effective band contains 301 points over 350--650 nm; weights include the calibrated response and photon conversion $\lambda/(hc)$, with the 650 nm point retaining a 1 nm quadrature width. Each training update samples 256 valid rays, cycling cameras every 16 updates. Training uses 128 equal-length segments with stratified random queries; RGB evaluation uses 256 midpoint queries. Image strides are two for synthetic data and four for measured data, giving rectified sizes of $192\times384$ and $612\times512$ (width by height). Volume priors use 1,024 uniform box samples. After update 12,000, the mean-soot term uses 1,024 separate ROI samples with volume-fraction correction.

The axis-aligned domain for synthetic data has $x,y\in[-0.24613505,0.24613505]$ and $z\in[0.00173265,1.59653470]$ m. For measured data, the bounds are $x\in[-0.039,0.047]$, $y\in[-0.042,0.029]$, and $z\in[-0.002,0.286]$ m.

\noindent\textbf{Spatial and chromaticity constraints.}
Foreground extraction uses maximum valid-channel signal with low/high thresholds of 5/10 DN equivalent for synthetic data and 10/30 DN for dark-corrected camera images. Eight-connected components require at least eight high-threshold seed pixels; a $3\times3$ closing is applied. For synthetic data, protection margins and transition width are zero. For measured data, both fields use 16-native-pixel margins, with an 8-native-pixel linear transition. Exterior weights use the third-largest valid view contribution; at least three valid views are required, and invalid views abstain. Support uses training images only.

Both datasets apply fixed soot exclusion in confidently exterior regions and a soft 300 K temperature prior. Soot is excluded only when all interpolation neighbors have full exterior weight and sufficient valid views. Chromaticity pixels require every channel to be at least 1 DN equivalent and the maximum channel at least 10 DN equivalent. Saturation limits are 4095 DN equivalent for synthetic data and 4094 DN before dark subtraction for measured data; all channels must be valid. The chromaticity denominator has a 1 DN floor.

\subsubsection{Appearance Reconstruction Baselines}
\label{sec:supp_appearance_implementation}

We adapt the official implementations pinned at \href{https://github.com/NVlabs/instant-ngp/tree/abe236ee00cf90cfca6e36e65c00435d5b21f50a}{Instant-NGP (\texttt{abe236e})}, \href{https://github.com/apchenstu/TensoRF/tree/6a1e23b10b9521221049aec843ab370a747a7982}{TensoRF (\texttt{6a1e23b})}, and \href{https://github.com/graphdeco-inria/gaussian-splatting/tree/54c035f7834b564019656c3e3fcc3646292f727d}{3DGS (\texttt{54c035f})}. All receive the same calibrated linear RGB, valid-pixel masks, rectified intrinsics, and fixed poses, with black backgrounds and no exposure optimization. Camera conventions and scene normalization follow each upstream implementation.

\noindent\textbf{Instant-NGP.}
We use the official NeRF \texttt{base.json}, with eight hash levels, four features per entry, finest target resolution 2048, and \texttt{aabb\_scale=1}. Training uses seed 1337, Huber loss, and Adam with initial learning rate $10^{-2}$ and $\beta=(0.9,0.99)$. Learning rate is multiplied by 0.33 every 10,000 steps from step 20,000. Linear-color training and pixel-center rays are enabled; random backgrounds and camera optimization are disabled.

\noindent\textbf{TensoRF.}
\texttt{TensorVMSplit} uses density/appearance ranks $(16,16,16)/(48,48,48)$, 27-dimensional appearance features, and an \texttt{MLP\_Fea} decoder. Batches contain 4096 valid pixels, with seed 20211202. The voxel budget grows from 2,097,156 to 27,000,000, with upsampling after updates 2,001, 3,001, 4,001, 5,501, and 7,001. Factor/decoder learning rates reset to 0.02/0.001 at upsampling. Alpha masks refresh after updates 2,001 and 4,001 at threshold $10^{-4}$, with initial box shrinking. Ray step ratio is 0.5. Training uses linear-RGB MSE with zero density $\ell_1$, TV, and rendering-weight cutoff.

\noindent\textbf{3DGS.}
We initialize 10,000 uniformly sampled points (point-cloud seed 42; training seed zero). Spherical-harmonics degree starts at zero and increases by one every 1,000 iterations, up to three. Densification runs every 100 iterations in the 500--15,000 window with gradient threshold $2\times10^{-4}$; opacity pruning uses 0.005. Opacity is capped at 0.01 every 3,000 iterations before densification ends. Training uses valid-pixel linear-RGB $\ell_1$ loss, zero DSSIM/depth-loss weights, and upstream Adam settings. Renderer outputs are clamped to $[0,1]$.

\subsubsection{Voxel and Voxel+TV}
\label{sec:supp_voxel_tv_implementation}

Voxel replaces the neural field with an $80\times80\times256$ node grid while retaining ThermoSight's renderer, calibration, full loss weights, activation schedule, and sampling. Temperature and soot logits are initialized from $\mathcal N(0,0.01^2)$ with seed 42 and use the same physical output maps, giving initial values near 1300 K and 0.347 ppm. Adam uses $\beta=(0.9,0.999)$ and the main model's learning-rate schedule. Trilinear interpolation operates on physical nodal values, with nodes aligned to box boundaries and border values outside the grid; soot exclusion follows interpolation.

Voxel+TV adds a weight-$10^{-7}$ anisotropic TV penalty; Voxel uses zero TV weight. With $q_T=(T-300\,\mathrm K)/(2000\,\mathrm K)$ and $q_f=f_v/10^{-6}$, the penalty is
\begin{equation}
    \mathcal L_{\mathrm{TV}}
    =\frac{1}{6}\sum_{d\in\{x,y,z\}}\sum_{q\in\{q_T,q_f\}}
    \operatorname{mean}_{\mathcal E_d}\frac{|\Delta_d q|}{h_d},
\end{equation}
where $\mathcal E_d$ denotes adjacent node pairs and $h_d$ their spacing in metres. TV uses the nodal fields before soot exclusion.

\subsubsection{Two-stage Tomography Baselines}
\label{sec:supp_tomography_implementation}

SIRT and ASD-POCS use an $80\times80\times256$ voxel-cell grid and an optically thin Siddon projector. Ray--voxel intersection lengths are in metres, and backprojection uses the matrix transpose. Each RGB channel starts from zero and uses the nine input views with nonnegative projection. SIRT runs 500 iterations with relaxation 1.0 and row/column normalization. ASD-POCS runs 500 camera-block SART outer iterations, each with at most 20 spacing-aware isotropic-TV updates. Its data step starts at 1.0 and decays by 0.99 per outer iteration; the initial TV step is 0.002 times the data-update displacement, with adaptive backtracking. Both datasets use the same settings.

The subsequent regularized pyrometry fits reconstructed emission $\mathbf j_i$ to the unit-soot spectral response $\mathbf b(T_i)$:
\begin{equation}
\begin{aligned}
    \mathcal E&=\sum_{i\in V}\ell_i
    +\frac{\mu}{6}\sum_{\{i,j\}\in E}a_{d(i,j)}
    \left(\frac{T_i-T_j}{2000\,\mathrm K}\right)^2,\\
    \ell_i&=\frac{\|\mathbf j_i-f_i\mathbf b(T_i)\|_2^2}
    {\|\mathbf j_i\|_2^2}
    +\lambda\left(\frac{f_i}{f_{\mathrm{ref}}}\right)^2.
\end{aligned}
\end{equation}
Here, $f_i$ is soot volume fraction and $E$ contains valid axis-adjacent pairs, counted once. We use $\lambda=10^{-6}$, $f_{\mathrm{ref}}=3\times10^{-6}$, $\mu=0.1$, and $a_d=(h_{\min}/\Delta_d)^2$, where $h_{\min}$ is the shortest box half-length and $\Delta_d$ the cell spacing. Temperatures span 300--2300 K in 2 K increments, with bounded $0\leq f_i\leq1$ solved analytically at each candidate. Voxel-wise initialization is followed by 40 red--black coordinate-descent sweeps. The valid set $V$ requires $\sum_c j_{i,c}>10^{-4}S/L_{\mathrm{diag}}$, using the fixed response scale $S$ and box diagonal $L_{\mathrm{diag}}$.

\subsubsection{Optimization and Comparison Protocol}
\label{sec:supp_comparison_protocol}

Main comparisons use nine input views; all RGB evaluation uses held-out Cam08. Appearance and voxel baselines share ThermoSight's 30,000-iteration budget; 3DGS performs 29,999 parameter updates. Per-iteration computation differs across methods. Each method--dataset result reports one run and its final budgeted checkpoint; seeds are specified above. SIRT/ASD-POCS are deterministic, zero-initialized solves.

\subsubsection{Metric Aggregation}
\label{sec:supp_metric_aggregation}

RGB metrics pool calibrated linear predictions across valid pixels and channels of held-out Cam08 on common rectified grids. Physical-field methods are assessed through RGB reprojections. Fixed emitting-region masks require an observed maximum channel signal of at least 10 DN equivalent, using dark-corrected camera signals. Display mapping, illustrative crops, and observed-background compositing are excluded.

For reference samples $y_i$, predictions $\widehat y_i$, and $N$ valid scalar samples,
\begin{equation}
\begin{aligned}
    \operatorname{RMSE}&=\sqrt{\frac{1}{N}\sum_{i=1}^{N}(\widehat y_i-y_i)^2},\\
    e&=100\sqrt{\frac{\sum_{i=1}^{N}(\widehat y_i-y_i)^2}{\sum_{i=1}^{N}y_i^2}}.
\end{aligned}
\end{equation}
PSNR is $20\log_{10}(P/\operatorname{RMSE})$. Main comparisons use the maximum valid linear RGB channel value in the training views as one shared peak for all methods and both regions. Normalizing observations and predictions by this peak gives $P=1$; squared errors are pooled before computing PSNR. Sparse-view RGB evaluation retains the same Cam08 masks and nine-view reference peak.

Synthetic fields use a common $80\times80\times256$ voxel-center grid with equal voxel weights. Reference validity requires all eight CFD interpolation corners to be valid; the hot, sooty mask additionally requires $T\geq1000$ K and $f_v\geq0.01$ ppm. Invalid tomography inversions retain $T=300$ K and $f_v=0$ in the fixed-mask error calculation. RMSE units are K and ppm. Evaluation on measured data uses RGB agreement and field visualizations.

\subsection{Spectral Rendering and Temperature--Soot Ambiguity}
\label{sec:supp_spectral_derivations}

\noindent\textbf{Analytic ray-segment integration.}
Let $u$ be the distance from the near edge of segment $i$. Within the segment, $\kappa_{\lambda,i}$ and $B_\lambda(T_i)$ are held constant. Its contribution to the camera radiance is
\begin{equation}
\begin{aligned}
    \Delta L_{\lambda,i}
     &= \mathcal T_{\lambda,i}B_\lambda(T_i)
        \int_0^{\Delta s_i}\kappa_{\lambda,i}
                    e^{-\kappa_{\lambda,i}u}\,\mathrm du \\
     &= \mathcal T_{\lambda,i}
        \bigl(1-e^{-\kappa_{\lambda,i}\Delta s_i}\bigr)B_\lambda(T_i).
\end{aligned}
\end{equation}
Here $\mathcal T_{\lambda,i}$ accounts for the segments between $i$ and the camera. Summing over segments yields the main-paper rendering equation. For $\kappa_{\lambda,i}\Delta s_i\ll 1$, the in-segment emission factor reduces to $j_{\lambda,i}\Delta s_i$ to first order, as in the overview figure.

\noindent\textbf{What chromaticity can constrain.}
Write $a_\lambda=6\pi E(\lambda)/\lambda$, so $\kappa_\lambda=a_\lambda f_v$. For a single camera, suppress the camera index on its response $R_c$. If an entire ray is optically thin over the effective passband and nearly isothermal at $T_0$, its channel integral before the DN mapping satisfies
\begin{equation}
\begin{aligned}
    C_c &\simeq F\, b_c(T_0),\\
    F &= \int_{s_n}^{s_f} f_v(\mathbf r(s))\,\mathrm ds,\\
    b_c(T) &= \int_\Lambda R_c(\lambda)\,a_\lambda
                             B_\lambda(T)\,\mathrm d\lambda .
\end{aligned}
\end{equation}
In one channel, increasing $T_0$ or increasing the soot column $F$ can both make the pixel brighter. For nonzero signal, normalizing by the sum over channels cancels $F$:
\begin{equation}
    \frac{C_c}{\sum_{c'} C_{c'}}
      \simeq \frac{b_c(T_0)}{\sum_{c'} b_{c'}(T_0)}.
\end{equation}
The normalized signal can thus respond to the temperature-dependent shape of the Planck spectrum without relying on absolute brightness. With a fixed linear camera gain the same cancellation holds in dark-corrected DN, with each gain absorbed into $b_c$. For real rays, temperatures vary along the line of sight, self-absorption depends on soot and wavelength, and the measured cameras use a quadratic DN mapping. We therefore match chromaticity after the full forward model, where it complements intensity fitting without guaranteeing a unique field reconstruction.

\noindent\textbf{Why three attenuation coefficients are an approximation.}
Consider only the contribution of a rear emitting layer, with spectrum $S_\lambda\geq0$, after passing through a foreground soot column $F_{\mathrm{fg}}$. Its channel signal relative to the same emitter without the foreground attenuation is
\begin{equation}
\begin{aligned}
    A_c(F_{\mathrm{fg}}) &=
      \int_\Lambda w_c(\lambda)e^{-a_\lambda F_{\mathrm{fg}}}
                         \,\mathrm d\lambda,\\
    w_c(\lambda) &=
      \frac{R_c(\lambda)S_\lambda}
           {\int_\Lambda R_c(\ell)S_\ell\,\mathrm d\ell}.
\end{aligned}
\end{equation}
For a given positive $F_{\mathrm{fg}}$, the equivalent coefficient is
$-\log A_c(F_{\mathrm{fg}})/F_{\mathrm{fg}}$. Its weights depend on the emitted spectrum, which changes with temperature. Even for a fixed spectrum, a single coefficient cannot reproduce $A_c$ for every column if $a_\lambda$ varies over wavelengths carrying signal: matching the first derivative at zero requires the weighted mean of $a_\lambda$, whereas matching the second derivative requires the mean of $a_\lambda^2$ to equal the square of that mean. These coincide only when the weighted variance vanishes. The three-channel transport ablation retains response-weighted emission but approximates spectral attenuation with one coefficient per channel. This distinction helps explain why nearly identical RGB scores can accompany different temperature and soot errors in the ablation.

\section{Convergence Analysis}
\label{sec:supp_convergence}
\begin{figure*}[!tbp]
    \centering
    \includegraphics[width=0.8\textwidth]{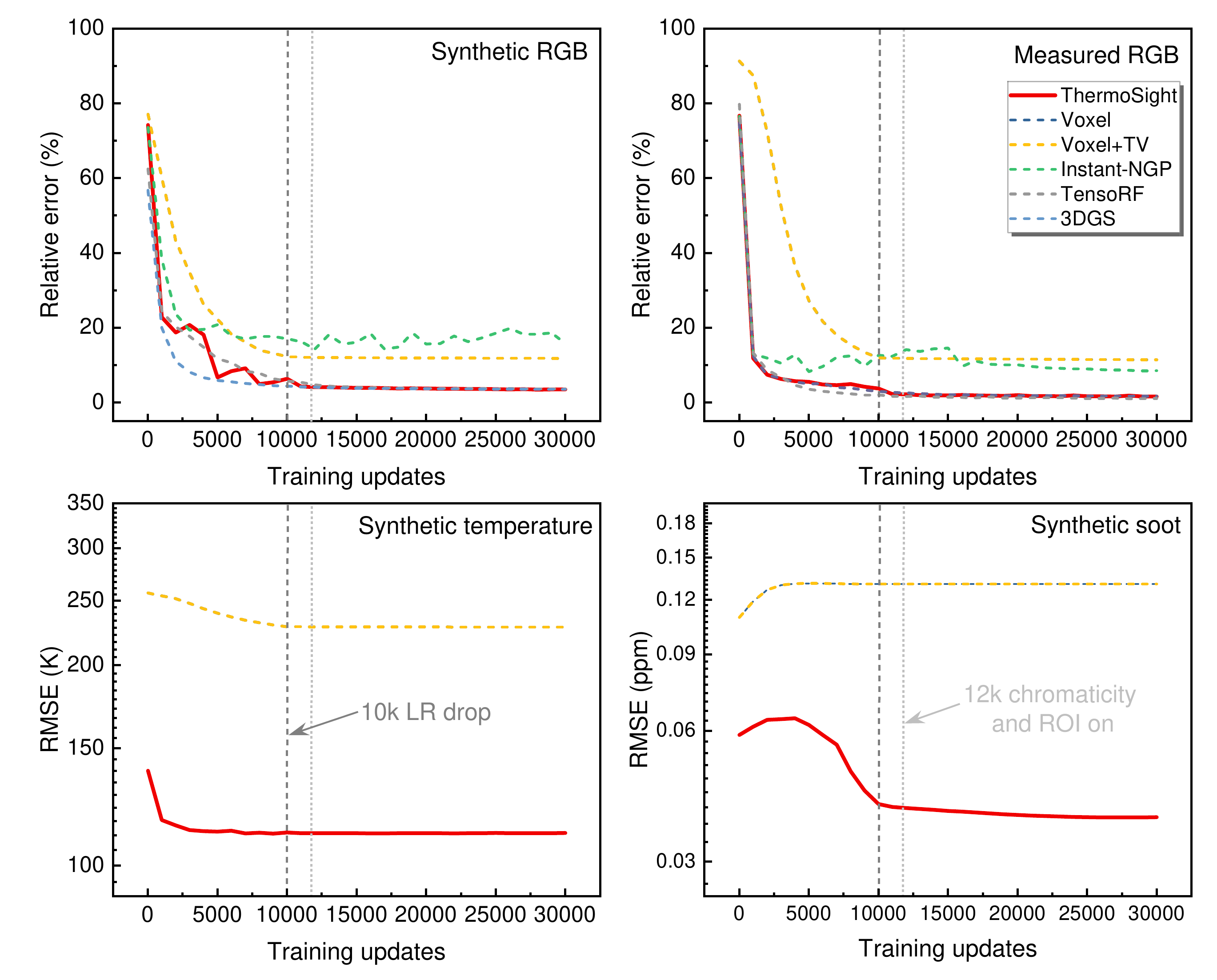}
    \captionsetup{skip=4pt}
    \caption{\textbf{Training convergence.} Top: RGB relative errors on held-out Cam08 for synthetic (left) and measured (right) data. Bottom: synthetic temperature RMSE (left) and soot RMSE (right) in the hot, sooty region. Curves compare checkpoints over the 30,000-update training budget. Vertical lines mark the learning-rate drop at 10,000 updates and activation of chromaticity supervision and ROI-based mean-soot sampling at 12,000 updates for ThermoSight and the voxel baselines. Voxel and Voxel+TV curves largely overlap.}
    \label{fig:supp_convergence}
\end{figure*}

\Cref{fig:supp_convergence} complements the final-checkpoint comparisons with training trajectories. ThermoSight's RGB errors decrease rapidly on both datasets and stabilize at levels close to the strongest appearance baselines. On synthetic data, its hot-region temperature error stabilizes early, while soot error continues to decrease before flattening in the later updates. Both physical-field errors remain below those of Voxel and Voxel+TV, whose nearly overlapping curves also show little late-stage change. This behavior supports the stability of the reported results within the shared training budget and confirms the limited effect of the added TV penalty under these settings.

\section{Sensitivity to Model Mismatch}
\label{sec:supp_mismatch}
\begin{figure*}[!tbp]
    \centering
    \includegraphics[width=0.9\textwidth]{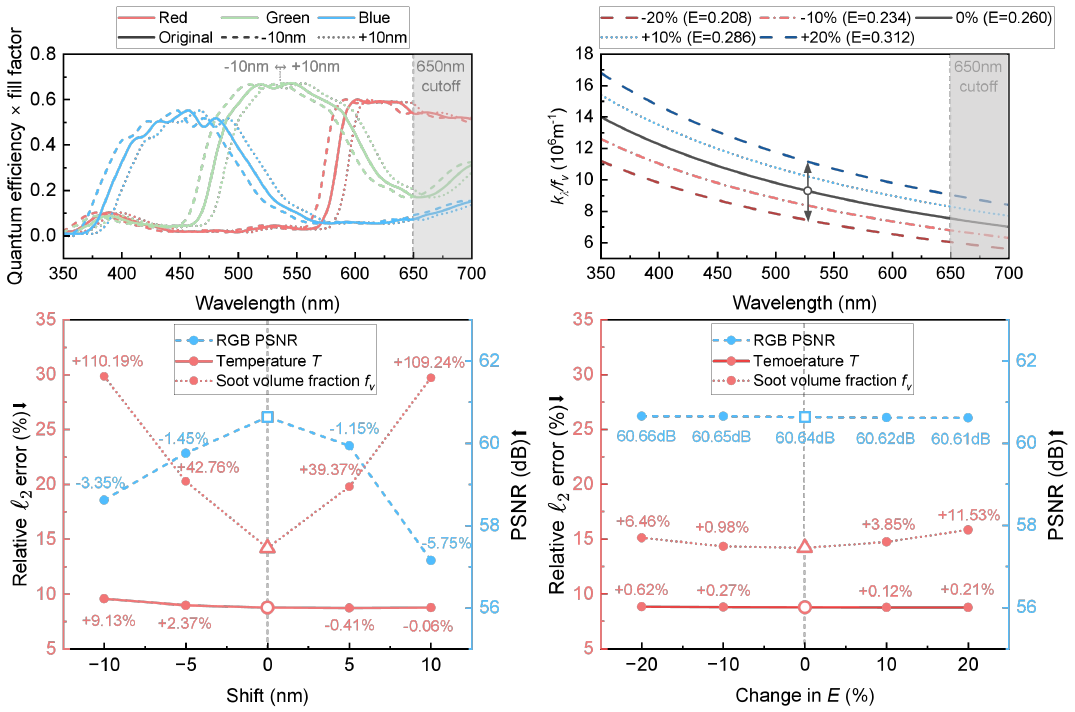}
    \captionsetup{skip=4pt}
    \caption{\textbf{Sensitivity to model mismatch on synthetic data.} Left: shifts in RGB channel spectral sensitivity; right: changes in the soot absorption function $E$. Top panels illustrate the perturbed spectral quantities, with the 650 nm cutoff marked. Bottom panels report temperature and soot relative $\ell_2$ errors in the hot, sooty region (left axes) and emitting-region RGB PSNR on held-out Cam08 (right axes). Annotations give each metric's relative change from the zero-mismatch baseline; PSNR itself is measured in dB.}
    \label{fig:supp_mismatch}
\end{figure*}

\noindent\textbf{Protocol.}
We keep the synthetic observations and CFD references fixed, perturb only the reconstruction model, and train each configuration from scratch for 30,000 updates using nine input views. Channel spectral-sensitivity curves are shifted by $\Delta\lambda\in\{-10,-5,0,5,10\}$ nm, with the filter cutoff unchanged. Separately, we set $E=0.26(1+\delta)$ for $\delta\in\{-20,-10,0,10,20\}\%$. The two tests share the matched-model baseline and otherwise retain the main comparison settings. These controlled perturbations assess sensitivity rather than represent measured calibration uncertainties.

\noindent\textbf{Results and interpretation.}
As shown in \cref{fig:supp_mismatch}, spectral shifts affect soot recovery more strongly than temperature recovery. At $\pm10$ nm, soot relative error increases by approximately 109--110\% over the matched case; temperature error increases by 9.13\% at $-10$ nm and is nearly unchanged at $+10$ nm. RGB PSNR decreases in both directions. The unequal field responses are consistent with changes in channel spectral weighting altering the temperature--soot trade-off.

Absorption-function amplitude changes have a different effect: RGB PSNR stays within 60.61--60.66 dB, while temperature and soot errors increase by up to 0.62\% and 11.53\%, respectively, relative to the matched case. The forward model has an exact scale ambiguity: for any $a>0$, replacing $E(\lambda)$ by $aE(\lambda)$ and $f_v(\mathbf x)$ by $f_v(\mathbf x)/a$ at fixed temperature leaves $\kappa_\lambda$ and $j_\lambda$ unchanged. All rendered camera signals are therefore identical under this transformation. This invariance applies to the image formation model; soot priors need not be invariant. Thus, strong image agreement alone cannot establish the absolute soot scale when the amplitude of $E$ is uncertain. These results characterize the dependence of quantitative reconstruction on spectral calibration and the prescribed soot optical properties.

\par\raggedbottom

\end{document}